\documentclass{article}

\PassOptionsToPackage{numbers,sort&compress}{natbib}

\usepackage[preprint]{neurips_2026}

\usepackage[utf8]{inputenc}
\usepackage[T1]{fontenc}
\usepackage{hyperref}
\usepackage{url}
\usepackage{booktabs}
\usepackage{array}
\usepackage{amsfonts}
\usepackage{amsmath}
\usepackage{graphicx}
\usepackage{microtype}
\usepackage{xcolor}
\usepackage{pifont}

\newcommand{\cmark}{\ding{51}}

\title{StereoGenBench: A Synthetic Multi-Camera Benchmark for Stereo Generation under Controlled Baseline Regimes}

\author{
  Yangzhi Cui\thanks{Equal contribution.\quad $^{\dagger}$Corresponding author.} \quad
  Feng Qiao$^{*}$ \quad
  Nathan Jacobs$^{\dagger}$ \\
  Washington University in St. Louis
}

\begin{document}
\maketitle

\begin{abstract}
Stereo image and video generation, stereo geometry estimation, and
condition-controlled view synthesis require paired data in which the variables
that determine binocular geometry---camera baseline, intrinsics, scene depth,
and camera motion---are known and controllable. Existing stereo resources
provide subsets of these variables, but resources commonly used for stereo
generation evaluation do not, to our knowledge, provide scene-paired,
calibrated multi-baseline right-view ground truth with jointly recorded
intrinsics, dense metric depth, and per-frame poses in a single controlled
source. We introduce \textbf{StereoGenBench}, a synthetic Unreal Engine benchmark
designed to make baseline-regime sensitivity and target-camera consistency
measurable under matched scene content. Each scene is rendered with a rigid six-camera lateral
array, yielding up to $\binom{6}{2}=15$ calibrated view pairs; adjacent baselines are sampled from inter-pupillary to wide-baseline regimes; focal
length is sampled independently; and every view is released with RGB, metric
depth, intrinsics, per-pair baselines, and per-frame poses. The splits include
two evaluation families for narrow and wide baseline regimes and a train-only
family for broader all-pairs coverage. We release the dataset, evaluation code,
reference results, Croissant metadata, and generation code/configuration for
extension with compatible assets. The dataset is available at
\url{https://huggingface.co/datasets/stereo-dataset/stereo-dataset}.
\end{abstract}

\section{Introduction}
\label{sec:intro}

Stereo image and video generation, stereo geometry estimation, and
condition-controlled view synthesis require evaluation resources that expose the
camera and scene variables governing binocular geometry. Real-world stereo
benchmarks fix these variables at capture time: KITTI~\citep{geiger2012kitti,
menze2015kitti} and DrivingStereo~\citep{yang2019drivingstereo} use
automotive-scale stereo rigs, Cityscapes~\citep{cordts2016cityscapes} uses a
fixed urban stereo rig, and Holopix50k~\citep{hua2020holopix50k} reflects the
short baseline of a mobile stereo camera. These resources are valuable, but
their baselines and intrinsics are largely fixed, metric depth is sparse or
absent in many cases, and camera motion is inherited from the capture platform.
Synthetic resources such as Scene Flow~\citep{mayer2016sceneflow},
SimStereo~\citep{jospin2022simstereo}, and
StereoCarla~\citep{guo2025stereocarla} provide dense rendered supervision, but
are still organized around dataset-specific rig configurations. To our
knowledge, no resource commonly used for stereo generation evaluation provides
scene-paired, calibrated multi-baseline right-view ground truth with jointly
recorded intrinsics, dense metric depth, and per-frame multi-camera poses in a
single controlled source.

This gap matters because the baseline is not a nuisance variable. For a rectified
stereo pair, disparity scales as
% \[
% d = \frac{B f_x}{z},
% \]
$d = B f_x z^{-1}$,
where \(B\) is the camera baseline, \(f_x\) is the pixel focal length, and \(z\)
is camera-frame depth. Training and evaluating only at a fixed baseline, therefore
conflates a method's stereo-scale behavior with the scene distribution on which it is tested. Current stereo image and video generation methods---including
warp-and-inpaint approaches~\citep{xie2016deep3d, shih2020_3dphoto},
mono-depth-driven pipelines~\citep{watson2020learning,
wang2025zerostereo, yu2025mono2stereo}, diffusion-based
generators~\citep{wang2024stereodiffusion, qiao2025genstereo,
liu2025dms, behrens2025stereospace}, and stereo video
synthesizers~\citep{dai2025svg, zhao2024stereocrafter,
shi2024immersepro, shvetsova2025m2svid}---often inherit implicit stereo-scale
priors from their training data or expose only method-specific stereo controls.
A benchmark with matched scene content and calibrated baseline variation can
separate several questions that are otherwise conflated: whether performance
changes with realized baseline, whether a generated right view matches the
specified target camera, and, for methods that expose a calibrated target-camera
interface, whether the method uses that interface rather than producing stereo
at a preferred implicit scale.

We introduce \textbf{StereoGenBench}, a synthetic Unreal Engine resource that
makes calibrated baseline response measurable under matched scene content. Each
scene is rendered with a rigid six-camera lateral array, admitting up to
$\binom{6}{2}=15$ calibrated view pairs from the same scene. Per-scene adjacent
baselines are sampled from inter-pupillary and wide-baseline regimes; focal
length and sensor dimensions are sampled and recorded; dense metric depth is
rendered for every camera; and per-frame six-camera poses are released alongside
RGB. The dataset is organized into two evaluation families,
\emph{IPD\_Gaussian} and \emph{Uniform}, and a train-only
\emph{Pairwise\_Uniform} family designed to improve all-pairs baseline coverage
when the six-camera rig is expanded into multiple training pairs.

The reference benchmark in this paper evaluates right-view generation for stereo
images and videos. We report results by inference condition rather than as a
single leaderboard. Methods in the oracle geometry-conditioned setting receive
target-view geometry such as ground-truth disparity, depth, or a ground-truth
warped right view; methods in the calibrated target-camera setting receive
camera metadata such as baseline or intrinsics but no ground-truth depth; and
unaligned monocular methods receive only the left view and their own implicit
geometry. These settings answer different questions, so cross-tier comparisons
should be read as diagnostic rather than as a unified ranking. The reference
results illustrate off-the-shelf behavior of representative public methods on
StereoGenBench and show how the benchmark surfaces geometric drift under
different baseline regimes.

\paragraph{Contributions.}
This work contributes: (i) a scene-paired synthetic stereo resource with a
six-camera calibrated rig, dense metric depth, intrinsics, per-pair baselines,
and per-frame poses; (ii) a split design that separates narrow IPD-scale
evaluation, wide-baseline evaluation, and train-only all-pairs baseline
coverage; (iii) a benchmark protocol that separates oracle geometry-conditioned,
calibrated target-camera, and unaligned monocular inference settings; and (iv)
public release of dataset metadata, evaluation code, reference results,
Croissant metadata, and the full generation pipeline, enabling users to create
compatible extensions with custom maps, assets, and camera settings.

% ============================================================
\section{Related Work}
\label{sec:related}

\paragraph{Stereo generation.}
Stereo generation methods synthesize the right-eye view of a stereo pair from a
monocular input, either for single images or temporally coherent video. Image
methods include warp-and-inpaint designs~\citep{xie2016deep3d, shih2020_3dphoto},
mono-depth-driven pipelines~\citep{watson2020learning, wang2025zerostereo,
yu2025mono2stereo}, and diffusion-based generators~\citep{wang2024stereodiffusion,
qiao2025genstereo, behrens2025stereospace, garg2025text2stereo}. Video methods
extend these ideas with temporal warping, stereo inpainting, video diffusion, or
feed-forward stereo conversion~\citep{dai2025svg, jin2024tsvg,
zhao2024stereocrafter, shi2024stereocrafter0, shi2024immersepro,
metzger2025elastic3d, shen2025stereopilot, xing2026stereoworld}. Some recent
methods expose explicit stereo controls: DMS~\citep{liu2025dms} uses directional
prompts to synthesize epipolar-aligned shifted and intermediate views, while
Elastic3D~\citep{metzger2025elastic3d} exposes a scalar control over stereo
effect strength or disparity range. These controls are useful, but they are
method-specific rather than calibrated metric baselines tied to a recorded target
camera.

\paragraph{Stereo datasets and benchmarks.}
Real-world stereo and multi-view resources such as KITTI~\citep{geiger2012kitti,
menze2015kitti}, DrivingStereo~\citep{yang2019drivingstereo},
Cityscapes~\citep{cordts2016cityscapes}, Holopix50k~\citep{hua2020holopix50k},
Middlebury~\citep{scharstein2014middlebury}, InStereo2K~\citep{bao2020instereo2k},
and ETH3D~\citep{schoeps2017eth3d} provide real captures with calibrated
geometry or ground truth, but are not designed to expose calibrated baseline as a
primary per-scene control variable for stereo generation. Synthetic resources
such as Scene Flow~\citep{mayer2016sceneflow}, MPI Sintel~\citep{butler2012sintel},
Virtual KITTI 2~\citep{cabon2020vkitti2}, TartanAir~\citep{wang2020tartanair},
IRS~\citep{wang2021irs}, SimStereo~\citep{jospin2022simstereo},
Dynamic Replica~\citep{karaev2023dynamicstereo}, Spring~\citep{mehl2023spring},
and StereoCarla~\citep{guo2025stereocarla} provide rendered supervision, stereo
videos, or diverse camera configurations. StereoCarla is especially relevant
because it includes diverse baselines and sensor placements for autonomous-driving
stereo matching, but it is not designed as a scene-paired multi-baseline
right-view generation benchmark with six-camera all-pairs metadata. Internet-
derived resources such as Stereo4D~\citep{jin2025stereo4d} mine stereo videos
to construct large-scale pseudo-metric 4D reconstructions, but likewise do not
provide controlled metric baseline sweeps under matched synthetic scene content.

\paragraph{Positioning.}
StereoGenBench is complementary to these resources. We do not aim to replace
real fixed-rig stereo benchmarks, which remain essential for measuring real-world
performance. Instead, StereoGenBench isolates a geometry variable that existing
resources do not make central for stereo generation evaluation: how a method
responds when matched scene content is rendered and evaluated under different
calibrated baseline regimes. Its six-camera rig, independently sampled intrinsics,
dense metric depth, and per-frame poses support baseline-stratified stereo
generation, all-pairs view synthesis, and disparity-scale diagnostics under
controlled synthetic conditions.

%---------------------------------------------------------------------------------------
\section{Dataset and Simulation Method}
\label{sec:dataset}

StereoGenBench differs from existing stereo resources in a single
structural choice: the variables that determine binocular geometry---
camera baseline, camera intrinsics, scene depth, and camera
trajectory---are released as per-scene controlled variables rather than
fixed at capture time. Real-world stereo benchmarks fix the rig
baseline at capture time and provide sparse or no metric depth;
existing synthetic resources provide dense depth but typically retain
a fixed rig and fixed intrinsics
(Table~\ref{tab:related-datasets}). StereoGenBench varies all four
axes within a single source while preserving paired ground truth,
which is the property that makes baseline-stratified and
intrinsics-stratified evaluation possible on the same scenes. The
remainder of this section describes the rig and baseline-sampling
regimes that realize the geometric axes
(Section~\ref{sec:dataset-rig-regimes}), the generation pipeline that
produces validated scenes (Section~\ref{sec:dataset-pipeline}), and
the current snapshot and quality control summary
(Section~\ref{sec:dataset-stats}).

\begin{table}[!h]
  \centering
  \scriptsize
  \setlength{\tabcolsep}{4pt}
  \caption{Stereo evaluation resources commonly used in stereo learning
  and stereo generation. ``Baseline'' is the rig configuration; ``Focal''
  reports whether per-scene focal length varies; ``Dense depth'' indicates
  per-pixel metric depth ground truth; ``Video'' indicates temporally
  coherent sequences; ``Multi-pair'' indicates multiple calibrated stereo
  pairs per scene. StereoGenBench is the only resource we are aware of
  designed for scene-paired, calibrated multi-baseline stereo generation
  evaluation with jointly recorded intrinsics, dense depth, and poses.}
  \label{tab:related-datasets}
  \begin{tabular}{l c c c c c}
    \toprule
    Resource & Baseline & Focal & Dense depth & Video & Multi-pair / scene \\
    \midrule
    \multicolumn{6}{l}{\emph{Real-world resources}} \\
    KITTI~\citep{geiger2012kitti, menze2015kitti}      & 54\,cm fixed       & fixed     & sparse & \cmark & --- \\
    Cityscapes~\citep{cordts2016cityscapes}            & 22\,cm fixed       & fixed     & sparse & \cmark & --- \\
    DrivingStereo~\citep{yang2019drivingstereo}        & 54\,cm fixed       & fixed     & sparse & \cmark & --- \\
    Holopix50k~\citep{hua2020holopix50k}               & $\sim$1.2\,cm      & fixed     & ---    & ---    & --- \\
    Middlebury 2014~\citep{scharstein2014middlebury}   & varying            & varying   & dense  & ---    & --- \\
    InStereo2K~\citep{bao2020instereo2k}               & 12\,cm fixed       & fixed     & dense  & ---    & --- \\
    ETH3D~\citep{schoeps2017eth3d}                     & varying            & varying   & dense  & ---    & --- \\
    MODEST~\citep{trivedi2025modest}                   & fixed              & 10\,$\times$\,5 grid & --- & --- & --- \\
    \midrule
    \multicolumn{6}{l}{\emph{Synthetic resources}} \\
    Scene Flow~\citep{mayer2016sceneflow}              & 1\,cm fixed        & fixed     & dense  & \cmark & --- \\
    MPI Sintel~\citep{butler2012sintel}                & 10\,cm fixed       & fixed     & dense  & \cmark & --- \\
    Virtual KITTI 2~\citep{cabon2020vkitti2}           & 53\,cm fixed       & fixed     & dense  & \cmark & --- \\
    TartanAir~\citep{wang2020tartanair}                & 25\,cm fixed       & fixed     & dense  & \cmark & --- \\
    IRS~\citep{wang2021irs}                            & 10\,cm fixed       & fixed     & dense  & \cmark & --- \\
    SimStereo~\citep{jospin2022simstereo}              & 16\,cm fixed       & fixed     & dense  & ---    & --- \\
    Dynamic Replica~\citep{karaev2023dynamicstereo}    & 4--30\,cm          & fixed     & dense  & \cmark & --- \\
    Infinigen SV~\citep{jing2024bidavideo}             & 7.5--120\,cm       & fixed     & dense  & \cmark & --- \\
    Spring~\citep{mehl2023spring}                      & 6.5\,cm fixed      & fixed     & dense  & \cmark & --- \\
    Stereo4D~\citep{jin2025stereo4d}                   & 6.3\,cm fixed      & fixed     & ---    & \cmark & --- \\
    StereoCarla~\citep{guo2025stereocarla}             & 5 fixed: 1--30\,cm & fixed     & dense  & \cmark & 5 \\
    \midrule
    \textbf{StereoGenBench (ours)} & \textbf{1--150\,cm; varying} & \textbf{18--85\,mm} & \textbf{dense} & \cmark & \textbf{15} \\
    \bottomrule
  \end{tabular}
\end{table}

\subsection{Six-Camera Rig and Baseline Regimes}
\label{sec:dataset-rig-regimes}

Each scene is rendered through six synchronized cameras forming a rigid,
rectified lateral array. Within each frame, all cameras share orientation and
differ only by lateral translation along a common local stereo axis, yielding up
to $\binom{6}{2}=15$ calibrated view pairs under shared scene content. Per-scene
focal length and sensor dimensions, adjacent spacings, all pairwise baselines,
and per-frame six-camera poses are recorded as metadata, allowing any selected
camera pair to be used for evaluation or analytic reference-disparity derivation.
\begin{figure}[!h]
  \centering
  \includegraphics[width=\linewidth]{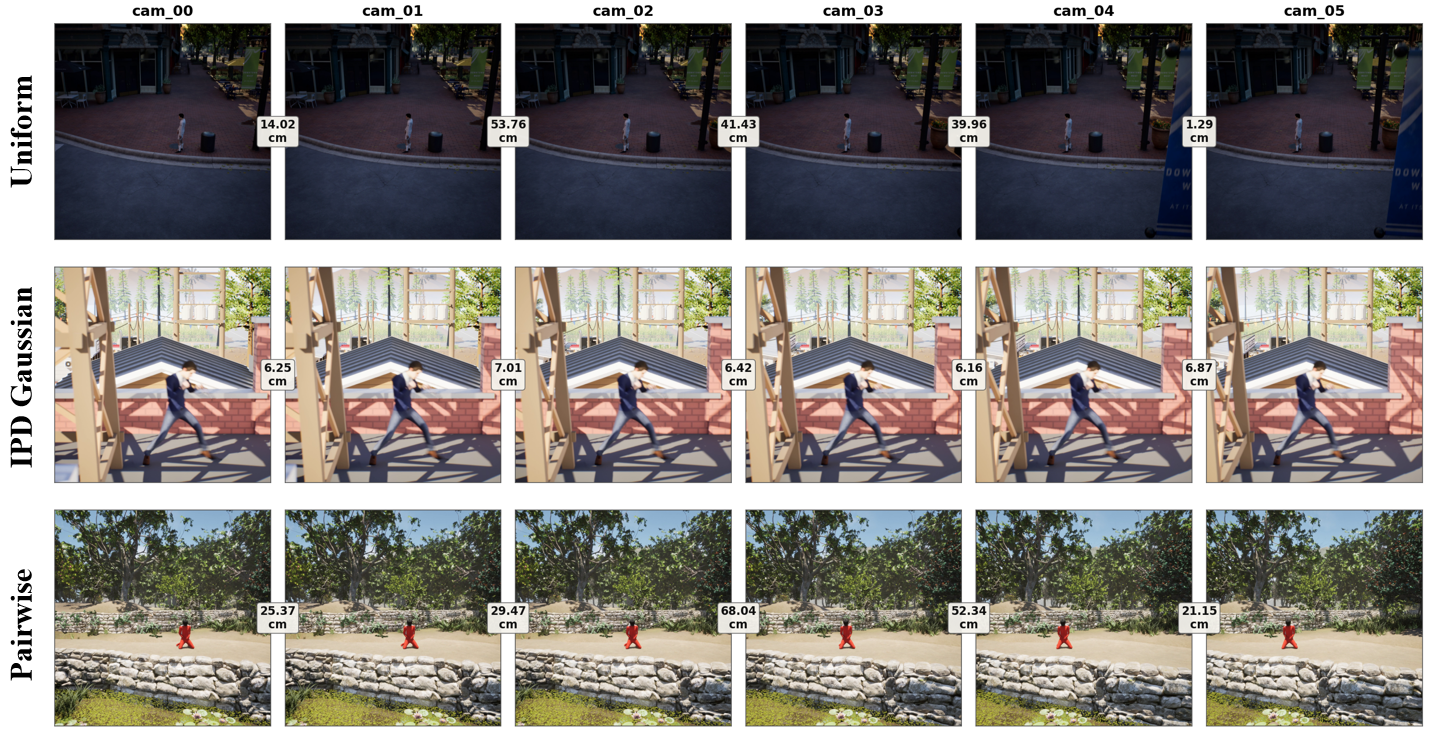}
  \caption{Six-camera rig outputs under the three baseline-sampling
  families. Each row shows six synchronized views from one scene at one
  time step; labels between adjacent views indicate realized camera
  spacings in centimeters.}
  \label{fig:rig}
\end{figure}

StereoGenBench contains three baseline-sampling families. \emph{IPD\_Gaussian}
samples adjacent spacings from a truncated Gaussian centered at $6.38$\,cm with
standard deviation $0.5$\,cm and clamp range $[4.5,8.5]$\,cm, reflecting an
inter-pupillary stereo regime. \emph{Uniform} samples adjacent spacings from
$\mathcal{U}[1.0,150.0]$\,cm and defines the wide-baseline evaluation branch.
Both families appear in train and eval. \emph{Pairwise\_Uniform} is train-only: it is designed to improve broad
all-pairs baseline coverage when each six-camera scene is expanded to all
$15$ rig pairs. The current
all-pairs distribution spans $5.11$--$200.00$\,cm. We report
\emph{Pairwise\_Uniform} realized adjacent statistics in
Table~\ref{tab:baseline-stats} and the realized all-pairs distribution in
Appendix~\ref{app:trainstats}.
\begin{table}[!h]
  \centering
  \small
  \caption{Realized adjacent-baseline statistics per split and family in
  the current snapshot. \emph{Pairwise\_Uniform} is reported for adjacent
  gaps only; its realized all-pairs distribution is reported in
  Appendix~\ref{app:trainstats}.}
  \label{tab:baseline-stats}
  \begin{tabular}{llrrrrr}
    \toprule
    Split & Family & \# adj.\ baselines & Min (cm) & Max (cm) & Mean (cm) & Std (cm) \\
    \midrule
    train & IPD\_Gaussian      & 16{,}445 & 4.5278 & 8.3867   & 6.3813  & 0.4998 \\
    train & Uniform            &  3{,}870 & 1.0113 & 149.8906 & 52.7908 & 38.9500 \\
    train & Pairwise\_Uniform  & 18{,}455 & 5.1117 & 123.2309 & 38.6091 & 20.7742 \\
    eval  & IPD\_Gaussian      &  1{,}780 & 4.7119 & 8.3332   & 6.3743  & 0.5052 \\
    eval  & Uniform            &  1{,}915 & 1.0058 & 149.7889 & 50.6741 & 39.1111 \\
    \bottomrule
  \end{tabular}
\end{table}

Each scene is a self-contained unit with six RGB videos, six metric-depth videos,
\texttt{baseline.json}, \texttt{trajectory.json}, and a completion marker. Stereo
disparity is derived from the recorded depth, relative pose, baseline, and
intrinsics rather than stored as a separate modality; file schemas are in
Appendix~\ref{app:schema}.

\subsection{Geometry Conventions and Sanity Checks}
\label{sec:geometry-validation}

Because StereoGenBench is intended as a geometry-sensitive benchmark, we
validate that the released depth streams, intrinsics, baselines, and camera
poses are numerically consistent with the evaluator's camera model. For an
image of width $W$ and height $H$, the evaluator derives pixel focal lengths
from the recorded physical focal length and sensor dimensions:
\begin{equation}
  f_x = \frac{f_{\mathrm{mm}}}{s_{w,\mathrm{mm}}} W,\qquad
  f_y = \frac{f_{\mathrm{mm}}}{s_{h,\mathrm{mm}}} H.
  \label{eq:intrinsics-conversion}
\end{equation}
For a rectified lateral pair with baseline $B$ and optical-axis camera depth
$z$, the reference disparity is
\begin{equation}
  d = \frac{B f_x}{z}.
  \label{eq:rectified-disparity}
\end{equation}
For non-adjacent pairs, the evaluator uses the full camera poses stored in
\texttt{trajectory.json}: source pixels are back-projected using source depth,
transformed by the recorded relative pose, and projected into the target camera
using the corresponding recorded intrinsics.

We validate on all complete scenes, $81$ frames. For each audited scene, we evaluated the
primary stereo pair (\texttt{cam\_00}$\rightarrow$\texttt{cam\_01}) and one long-baseline non-adjacent pair
(\texttt{cam\_00}$\rightarrow$\texttt{cam\_05}).  
Table~\ref{tab:geometry-validation-main} summarizes the validation results.
These checks test whether the recorded camera geometry is numerically
self-consistent and whether the released depth streams are coherent with the
recorded poses and intrinsics.
\begin{table}[!h]
  \centering
  \scriptsize
  \setlength{\tabcolsep}{4pt}
  \caption{Geometry consistency audit on StereoGenBench. The
  checks validate numerical consistency among released intrinsics, baselines,
  depth streams, and camera poses.}
  \label{tab:geometry-validation-main}
  \begin{tabular}{l p{5.6cm} c}
    \toprule
    Check & Quantity & Result \\
    \midrule
    Intrinsics &
    Maximum deviation between evaluator-derived $f_x,f_y$ and
    Eq.~\ref{eq:intrinsics-conversion}
    & $0$ px \\
    Pose / baseline &
    Maximum deviation between pose-derived baselines and the released adjacent
    and pairwise baseline fields
    & $6.68\times 10^{-12}$ cm \\
    Depth validity &
    Mean valid-depth ratio over audited frames and cameras
    & $94.59\%$ \\
    Reprojection &
    Median / 95th-pct left-to-right reprojection error on non-occluded pixels
    & $1.20\times 10^{-6}$ px / $4.39\times 10^{-3}$ px \\
    \bottomrule
  \end{tabular}
\end{table}

The audit indicates that the released intrinsics, baselines, and camera poses
are numerically self-consistent to near machine precision, and that the
evaluator's projection model is aligned with the released depth and pose
metadata. The strongest evidence comes from the pose/baseline and reprojection
checks: pose-derived baselines match the released baseline fields up to
rounding-level error, and non-occluded reprojection errors are far below one
pixel. 
\subsection{Generation Pipeline}
\label{sec:dataset-pipeline}

\begin{figure}[!h]
  \centering
  \includegraphics[width=\linewidth]{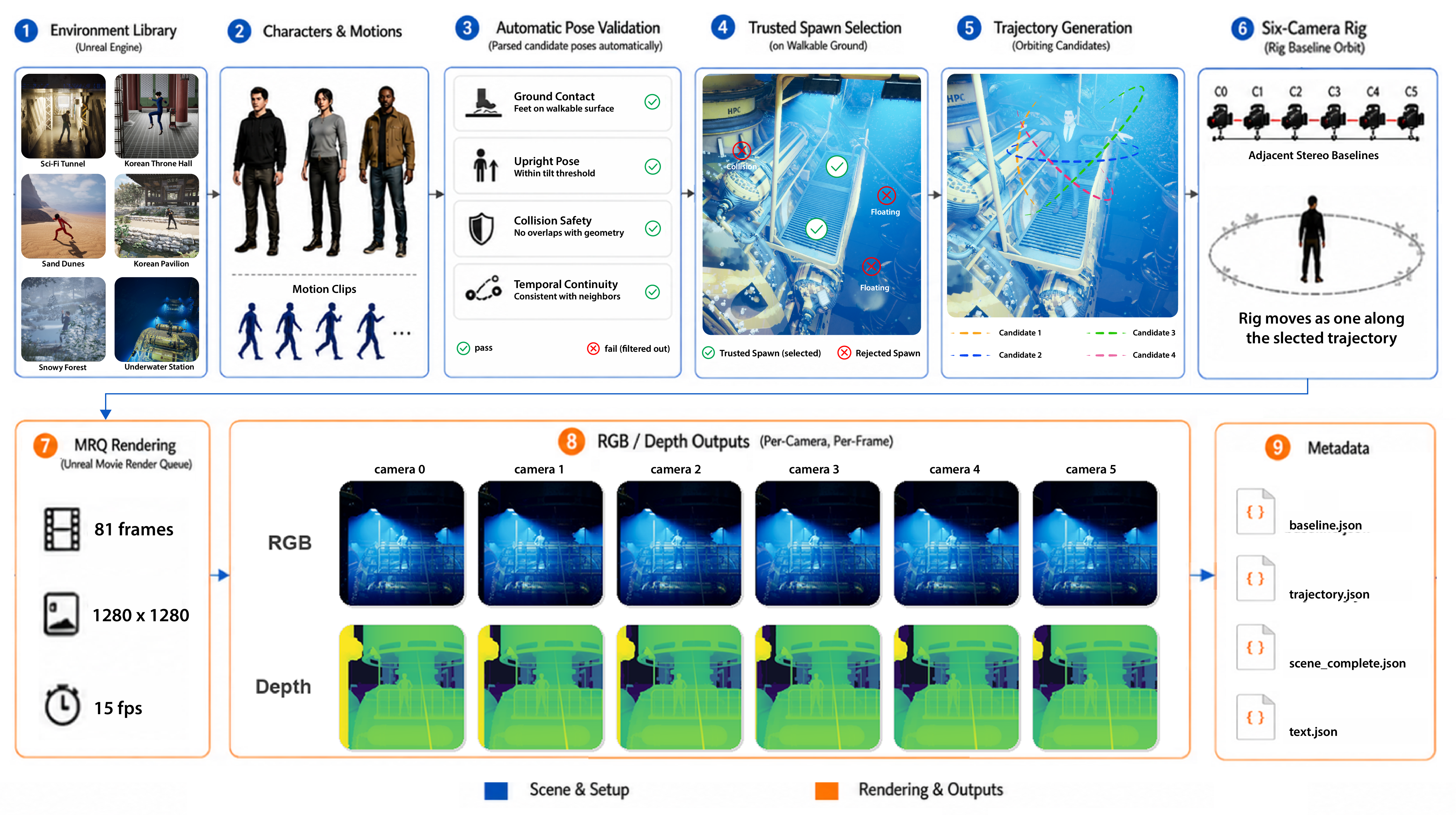}
  \caption{Generation pipeline. Scene construction, spawn validation,
  trajectory candidate ranking, six-camera rendering, output validation,
  and metadata export are performed before a scene is added to the released
  dataset.}
  \label{fig:pipeline}
\end{figure}

StereoGenBench is generated by an Unreal Engine Python pipeline driving Movie
Render Queue. For each scene, the pipeline proposes a map, character, animation,
spawn location, camera intrinsics, six-camera rig, and base trajectory; filters
the proposal for spawn support, subject visibility, camera collision,
trajectory continuity, and render validity; renders synchronized six-camera RGB
and depth sequences; and exports videos and metadata in the released scene
format. Figure~\ref{fig:pipeline} summarizes the generation workflow.

Scene construction is a proposal-and-filter process rather than an independent
uniform draw over all variables. As a result, the realized distributions in the
released dataset reflect both the proposal distributions and the validation
filters. The current snapshot draws from $25$ maps spanning indoor, outdoor,
urban, natural, sci-fi, and stylized environments, with $15$ character FBX files
and $30$ action FBX files. Camera trajectories are generated by sampling valid
viewpoints around the subject and connecting them into an $81$-frame path; when
a transition fails collision or visibility checks, the pipeline uses a
restart-and-splice procedure to complete a valid trajectory. Accepted scenes are
rendered as synchronized six-camera RGB and metric-depth sequences at
$1280\times1280$ resolution, $81$ frames, and $15$ fps, then compressed and
paired with \texttt{baseline.json}, \texttt{trajectory.json}, and a completion
marker. Detailed trajectory construction, pipeline parameters, and validation
thresholds are provided in Appendices~\ref{app:trajectory-construction}.

\subsection{Dataset Statistics and Quality Control}
\label{sec:dataset-stats}

StereoGenBench contains $8{,}493$ scenes: $7{,}754$ train scenes and
$739$ evaluation scenes. The evaluation split is further divided into
\emph{seen-map} and \emph{unseen-map} subsets. The seen-map subset is
scene-disjoint from training but uses maps that also appear in the training
split; it serves as the primary condition-controlled in-domain evaluation
subset. The unseen-map subset is map-disjoint from training and is reported
separately as a map-level generalization diagnostic. This separation is useful
because the public methods evaluated in this paper are used off-the-shelf rather
than retrained on StereoGenBench: seen--unseen gaps reflect map appearance and
visual-prior shifts, whereas \emph{IPD\_Gaussian}--\emph{Uniform} gaps probe
baseline-regime response. \emph{Pairwise\_Uniform} is train-only and does not
appear in evaluation. Scene, map, character, animation identifiers, and machine-generated scene-level
text descriptions are included in the metadata so users can construct
alternative held-out protocols and prompt- or retrieval-based evaluation
subsets. The text descriptions are intended as scene-level descriptive metadata,
not as human-verified semantic ground truth.
Representative scene examples are shown in Figure~\ref{fig:scene-diversity-main},
and per-map counts are reported in Appendix~\ref{app:map-counts}.
\begin{table}[!h]
  \centering
  \small
  \caption{Scene counts in StereoGenBench, by split and family.
  The evaluation split is separated into seen-map and unseen-map subsets.
  \emph{Pairwise\_Uniform} is a training-only family.}
  \label{tab:scene-counts}
  \begin{tabular}{lrrrr}
    \toprule
    Split & IPD\_Gaussian & Uniform & Pairwise\_Uniform & Total \\
    \midrule
    Train           & 3{,}289 &   774 & 3{,}691 & 7{,}754 \\
    Eval seen-map   &   265   &   281 & ---     &   546   \\
    Eval unseen-map &    91   &   102 & ---     &   193   \\
    \midrule
    Eval total      &   356   &   383 & ---     &   739   \\
    Total           & 3{,}645 & 1{,}157 & 3{,}691 & 8{,}493 \\
    \bottomrule
  \end{tabular}
\end{table}

\begin{figure}[!h]
  \centering
  \includegraphics[width=\linewidth]{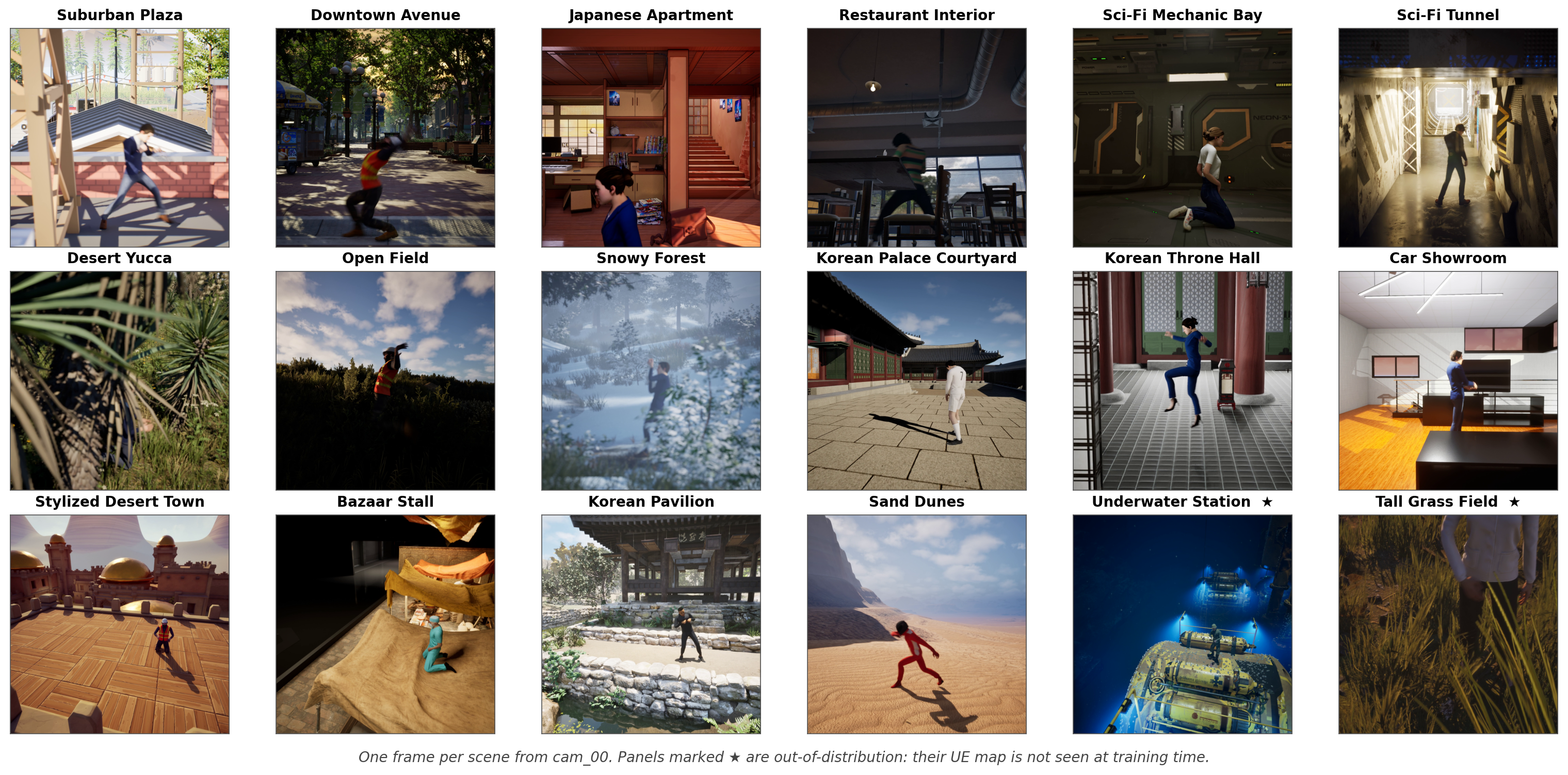}
  \caption{Representative scenes from the released map roots, illustrating
  indoor, outdoor, urban, natural, sci-fi, and stylized environments. Panels
  marked $\star$ (Underwater Station, Tall Grass Field) are out-of-distribution
  maps not seen at training time.}
  \label{fig:scene-diversity-main}
\end{figure}

Quality control combines automated filters with a preview-based manual pass.
Scene construction rejects invalid spawns, poses, animations, and trajectories;
post-render checks reject missing frames, invalid depth payloads, poor RGB
visibility, and edge-occlusion artifacts; a manual review pass inspected all rendered scene previews and moved
56 scene directories ($0.66\%$) into a filtering holdout; and a final audit verifies that each counted scene contains
the expected media and metadata files.

\section{Benchmark Protocol}
\label{sec:benchmark}

StereoGenBench supports several geometry-aware stereo tasks. The reference
benchmark in this paper focuses on right-view generation, but we report it
under three inference-condition tasks rather than as a single leaderboard,
because public methods expose different input contracts. \textbf{Task A / Tier
G0: oracle geometry-conditioned synthesis} evaluates methods that receive
\(I_L\) together with target-view information such as ground-truth disparity,
ground-truth depth, or a GT-derived warped right view. This task measures
synthesis, rendering, or inpainting quality under known target geometry; it is
not a monocular stereo-generation setting. \textbf{Task B / Tier G1:
calibrated target-camera generation} evaluates methods that receive \(I_L\) and
target-camera metadata such as baseline \(B\) and intrinsics \(K\), but no
ground-truth depth or disparity. \textbf{Task C / Tier G2: unaligned monocular
stress testing} evaluates methods that consume only \(I_L\) and their own
implicit monocular geometry, with no benchmark target-view information. The
tier labels in the result tables record the inference contract used for each
row and should be used to interpret, not rank, cross-tier results.

For image generation, a method maps a left RGB image \(I_L\) to a generated
right view \(\hat{I}_R\), which is scored against the rendered right view
\(I_R\) from the selected camera pair. For video generation, the same contract
is applied over the released \(T=81\)-frame sequences at \(15\) fps and
\(1280\times1280\) native resolution; metric computation uses the resized
resolution specified by the released evaluator. Reconstruction and stereo
diagnostics are computed frame-wise and then averaged over frames and scenes.
When FVD is reported, it is computed at the sequence level. Thus, the current
video results should be read as frame-wise stereo-generation diagnostics plus a
distributional video statistic; they do not by themselves isolate temporal
flicker or flow-warped temporal consistency.

Results are reported on two evaluation branches. \emph{IPD\_Gaussian} probes
narrow inter-pupillary-scale geometry, while \emph{Uniform} probes a wider
baseline range. The train-only \emph{Pairwise\_Uniform} family is not used in
the reference evaluation. Unless otherwise noted, each evaluated scene
contributes one primary stereo pair from the six-camera rig, with the realized
baseline read from \texttt{baseline.json}. The released metadata also supports
evaluation over all \(\binom{6}{2}=15\) view pairs for protocols requiring
within-scene baseline sweeps, but those sweeps are not part of the reference
snapshot.

We report three groups of metrics. \emph{Reconstruction quality} compares
\(\hat{I}_R\) with the rendered target \(I_R\) using PSNR,
SSIM~\citep{wang2004ssim}, and LPIPS~\citep{zhang2018lpips}. These metrics are
most directly meaningful for Task A, where target-view geometry is provided,
and should be interpreted as alignment diagnostics for Tasks B and C.

\emph{Stereo-geometry diagnostics} measure matchability and disparity-scale
fidelity. Let \(M_{gt}\) and \(M_{pred}\) denote the sets of left-image
keypoints that participate in accepted matches between \((I_L,I_R)\) and
\((I_L,\hat{I}_R)\), respectively, using the same DeDoDe
matcher~\citep{edstedt2024dedode}, keypoint budget, confidence threshold, and
epipolar tolerance. Two left-keypoint matches are counted as the same element
when their left-image coordinates fall within the evaluator tolerance. We
define
\begin{equation}
  \mathcal{E}_{\mathrm{Match}} = 100\left(1-
  \frac{|M_{gt}\cap M_{pred}|}{|M_{gt}\cup M_{pred}|}\right),
  \label{eq:ematch}
\end{equation}
which measures how much the set of stereo-matchable left-image structures
changes when the rendered right view is replaced by the generated right view.
P-PSNR is a target-free stereo-consistency diagnostic computed from \(I_L\) and
\(\hat{I}_R\) alone: for each source patch, the evaluator searches a horizontal
disparity window and records the PSNR of the best-matching patch. Because this
can reward local texture similarity even when global geometry is wrong, we
report P-PSNR only as a diagnostic. Finally, SD measures disparity-scale
fidelity. A fixed reference stereo matcher,
FoundationStereo~\citep{wen2025foundationstereo}, estimates disparity
\(\hat{D}\) from \((I_L,\hat{I}_R)\). Over finite valid pixels, after the outlier and occlusion filtering specified in the evaluator, we fit
\(D_{gt}\approx a\hat{D}+b\) and report $\mathrm{SD}=|a-1|$.
The released evaluator reports the fitting method, valid-pixel ratio, and
residual statistics in addition to SD.

\emph{Distributional metrics} compare generated and rendered target
distributions without requiring per-sample pixel alignment. FID is computed
between generated right views and rendered target right views over each
evaluation branch. FVD is computed between generated and rendered right-view
video sequences using an I3D feature extractor. For image-generation methods
evaluated on video scenes, we run the image model independently on each frame
and compute FVD on the resulting frame sequence; this measures distributional
sequence statistics and does not imply temporal modeling ability.

Absolute values of \(\mathcal{E}_{\mathrm{Match}}\), P-PSNR, SD, FID, and FVD
depend on the chosen reference systems, matcher budgets, feature extractors,
evaluation resolution, and valid-pixel filters. The exact configurations used
by the released evaluator are recorded in
Appendix~\ref{app:metric-calibration}, which also provides calibration controls
on rendered targets, copied-left views, wrong-baseline targets, and random right
views. These controls make the metric scales interpretable within
StereoGenBench by separating identity-target behavior, matchability without
stereo shift, plausible imagery with incorrect target-camera geometry, and
unmatched-scene behavior.

\section{Reference Results}
\label{sec:baselines}

Table~\ref{tab:main-results} reports the main reference snapshot on the
seen-map evaluation subset; map-disjoint unseen-map diagnostics are reported in
Appendix~\ref{app:unseen-results}. We include representative public stereo
generation methods that could be run end-to-end on StereoGenBench in the current
snapshot. Recent methods such as Elastic3D~\citep{metzger2025elastic3d},
Eye2Eye~\citep{geyer2025eye2eye}, and
StereoWorld~\citep{xing2026stereoworld} are excluded from the quantitative
measurement because public code or weights were unavailable at the submission date.

Rows are grouped by inference tier, following the protocol in
Section~\ref{sec:benchmark}. G0 rows receive target-view geometry, G1 rows
receive calibrated target-camera metadata, and G2 rows receive no benchmark
target-view information. Following the protocol in Section~\ref{sec:benchmark}, cross-tier
comparisons are diagnostic. Per-method interfaces, working resolutions, geometry
sources, and unavailable-cell classifications are recorded in Appendix~\ref{app:interfaces}.

The main pattern is that oracle target-geometry methods achieve stronger
pixel alignment, as expected from their input contract, while unaligned
monocular methods show larger disparity-scale drift because their outputs are
not anchored to the benchmark target camera. This effect is an implicit scale collapse measurable only because the
baseline is a controlled variable: in the baseline-stratified Uniform
breakdown (Appendix~\ref{app:baseline-bins}), the SD of the G2 method
Mono2Stereo rises monotonically from $2.63$ at $[1,10)$\,cm to $50.85$ at
$[100,150]$\,cm, whereas the G0 method GenStereo stays within
$0.03$--$0.09$ across the same bins. Across tiers, the
\emph{Uniform} branch is generally harder than \emph{IPD\_Gaussian}, reflecting
larger realized disparities and more disocclusion. High match-error values, including in some geometry-conditioned video rows,
should be interpreted as fixed-evaluator matchability diagnostics rather than
as standalone universal scores; the controls in
Appendix~\ref{app:metric-calibration} show how these diagnostics behave on
identity, copied-left, wrong-target-camera, and unmatched-scene inputs. 
\begin{table*}[!htbp]
  \centering
  \tiny
  \setlength{\tabcolsep}{5pt}
  \renewcommand{\arraystretch}{1.27}
  \caption{Reference results on StereoGenBench seen-map evaluation split. Methods
are grouped by inference tier (G0/G1/G2; see
Appendix~\ref{app:interfaces}); rows across tiers answer different questions
because they consume different inputs. Reconstruction metrics compare generated
and rendered right views; stereo diagnostics measure matchability and
disparity-scale fidelity; distributional metrics compare generated and rendered
target distributions. Reconstruction and stereo diagnostics are frame-wise; FVD
is sequence-level when reported.}
  \label{tab:main-results}
  \begin{tabular}{l c c l !{\vrule width 1pt} r r r !{\vrule width 1pt} r r r !{\vrule width 1pt} r r}
    \toprule
                                                                   &       &     &               & \multicolumn{3}{c!{\vrule width 1pt}}{Reconstruction} & \multicolumn{3}{c!{\vrule width 1pt}}{Geometric correctness} & \multicolumn{2}{c}{Distributional} \\
    Method & Type & Tier & Branch & PSNR$\uparrow$ & SSIM$\uparrow$ & LPIPS$\downarrow$ & $\mathcal{E}_\mathrm{Match}\downarrow$ & P-PSNR$\uparrow$ & SD$\downarrow$ & FID$\downarrow$ & FVD$\downarrow$ \\
    \midrule
    \multicolumn{12}{l}{\emph{Tier G0 --- target-view geometry}} \\
    GenStereo~\citep{qiao2025genstereo}                            & Image & G0 & Uniform       & 27.35 & 0.8271 & 0.1176 & 43.85 & 24.80 & 0.0471  &   7.97 & 138.57 \\
                                                                   &       &    & IPD\_Gaussian & 28.47 & 0.8618 & 0.0947 & 40.75 & 28.66 & 0.1039  &   4.40 &  31.67 \\
    StereoDiffusion~\citep{wang2024stereodiffusion}                & Image & G0 & Uniform       & 22.76 & 0.6952 & 0.1916 & 46.00 & 23.28 & 0.2163  &  20.44 & 337.20 \\
                                                                   &       &    & IPD\_Gaussian & 23.02 & 0.7236 & 0.1569 & 43.23 & 25.98 & 0.5124  &   6.79 &  90.34 \\
    ZeroStereo~\citep{wang2025zerostereo}                          & Image & G0 & Uniform       & 26.02 & 0.8608 & 0.1193 & 36.19 & 25.45 & 0.0299  &  12.65 & 293.89 \\
                                                                   &       &    & IPD\_Gaussian & 29.21 & 0.9058 & 0.0889 & 36.46 & 30.50 & 0.0355  &   4.81 &  48.60 \\
    Stereo-from-Mono~\citep{watson2020learning}                    & Image & G0 & Uniform       & 23.17 & 0.7615 & 0.1490 & 41.37 & 26.78 & 0.1150  &  20.12 &  75.55 \\
                                                                   &       &    & IPD\_Gaussian & 24.04 & 0.8410 & 0.0983 & 36.74 & 31.88 & 0.1709  &   6.08 &  22.21 \\
    SVG~\citep{dai2025svg}                                         & Video & G0 & Uniform       & 23.40 & 0.7322 & 0.2299 & 46.30 & 23.37 & 0.0969  &  30.08 & 105.06 \\
                                                                   &       &    & IPD\_Gaussian & 25.07 & 0.7745 & 0.1912 & 42.95 & 26.41 & 0.1329  &  19.36 &  71.93 \\
    StereoCrafter~\citep{zhao2024stereocrafter}                    & Video & G0 & Uniform       & 21.93 & 0.6580 & 0.3198 & 46.06 & 22.19 & 0.1698  &  44.66 & 462.82 \\
                                                                   &       &    & IPD\_Gaussian & 22.96 & 0.7170 & 0.2602 & 44.25 & 24.32 & 0.2779  &  25.13 & 106.85 \\
    \midrule
    \multicolumn{12}{l}{\emph{Tier G1 --- target-camera metadata}} \\
    StereoSpace~\citep{behrens2025stereospace}                     & Image & G1 & Uniform       & 19.71 & 0.5821 & 0.2425 & 46.60 & 24.08 & 0.3951  &  11.57 & 227.32 \\
                                                                   &       &    & IPD\_Gaussian & 18.43 & 0.6069 & 0.2435 & 41.56 & 24.79 & 0.5346  &   6.55 &  64.08 \\
    \midrule
    \multicolumn{12}{l}{\emph{Tier G2 --- unaligned monocular}} \\
    Mono2Stereo~\citep{yu2025mono2stereo}                          & Image & G2 & Uniform       & 18.30 & 0.5420 & 0.3015 & 52.48 & 25.22 & 18.8764 &  13.59 &  42.04 \\
                                                                   &       &    & IPD\_Gaussian & 17.89 & 0.5910 & 0.2469 & 46.72 & 24.94 &  6.6318 &   5.67 &  14.55 \\
    ImmersePro~\citep{shi2024immersepro}                           & Video & G2 & Uniform       & 18.08 & 0.5444 & 0.3311 & 50.86 & 24.09 &  1.9278 &  13.47 &  40.33 \\
                                                                   &       &    & IPD\_Gaussian & 17.66 & 0.6030 & 0.2824 & 45.92 & 23.59 &  0.9974 &   4.58 &  12.59 \\
    StereoCrafter-Zero~\citep{shi2024stereocrafter0}               & Video & G2 & Uniform       & 12.19 & 0.3935 & 0.6637 & 68.34 & 13.45 &  0.8390 & 135.43 & 896.43 \\
                                                                   &       &    & IPD\_Gaussian & 11.76 & 0.3978 & 0.6997 & 72.44 & 13.14 &  0.8962 & 108.01 & 991.66 \\
    StereoPilot~\citep{shen2025stereopilot}                        & Video & G2 & Uniform       & 14.67 & 0.4382 & 0.5353 & 48.36 & 16.99 &  0.8528 &  50.69 & 259.70 \\
                                                                   &       &    & IPD\_Gaussian & 13.89 & 0.4683 & 0.5434 & 59.43 & 15.67 &  0.8251 &  37.82 & 332.50 \\
    \bottomrule
  \end{tabular}
\end{table*}

\section{Limitations, Responsible Use, and Release}
\label{sec:limitations}
\label{sec:rai-hosting}

StereoGenBench is synthetic, and results obtained on it should not be treated as
direct evidence of real-world performance without a separate real-data
correlation study. Rendering assumptions, simulator-specific lighting and
materials, the finite Unreal asset pool, stylized map content, and heuristic
spawn / framing / animation validation introduce domain-specific biases. The
dataset contains synthetic human avatars drawn from a limited asset pool; it
should not be used to study demographic representativeness, real human behavior,
identity, biometrics, or surveillance.

Quality control combines automated filters, post-render validation, a
preview-based manual pass, and a final release audit. Passing these checks
indicates the absence of detected failures under our filters; it is not a
certificate that all six camera views are artifact-free, that simulated motion is
physically exact, or that all material depth values are unambiguous. Users
studying geometry at the pixel level should inspect the released validation
metadata and, where necessary, subselect scenes by map, depth validity,
reprojection-error statistics, or material type.

The benchmark metadata defines target camera geometry, but public method
wrappers consume different subsets of this information. We therefore report
results by inference tier rather than enforcing a single input contract such as
\((I_L,B)\) or \((I_L,B,K)\). The current video results are frame-wise sequence
diagnostics and do not by themselves measure temporal flicker or flow-warped
temporal consistency. The released metadata supports additional protocols beyond
this submission, including all-pairs sweeps over the six-camera rig,
intrinsics-stratified evaluation, cross-regime training, map-held-out subselects,
and sim-to-real rank-correlation studies.

StereoGenBench contains no real personal data. It is intended for evaluation and
development of stereo generation, stereo geometry, view synthesis, and depth
methods under controlled camera conditions. It is not intended for surveillance,
biometric identification, identity inference, demographic analysis, human
behavior recognition, or impersonation-oriented synthetic-media training.
Dataset license, Croissant metadata, RAI fields, anonymized endpoints, code
release, asset provenance, compute footprint, and exact-regeneration
requirements are documented in Appendix~\ref{app:hosting}. The rendered outputs
and metadata are distributed under the dataset license specified there; source
Unreal Engine maps, character meshes, and animation assets retain their original
licenses and are required only for exact regeneration or extension with the same
assets.

\section{Conclusion}
\label{sec:conclusion}

StereoGenBench makes calibrated baseline response and target-camera consistency
measurable under matched scene content. By rendering each scene with a
six-camera calibrated rig and releasing RGB, metric depth, intrinsics, baselines,
and per-frame poses, it supports right-view generation evaluation under controlled
narrow- and wide-baseline regimes. The reference results illustrate off-the-shelf
behavior across inference tiers and motivate task-separated reporting rather than
a unified leaderboard. The release enables future studies of baseline
conditioning, multi-pair stereo generation, stereo matching, depth estimation,
and sim-to-real correlation.

\clearpage

\bibliographystyle{plainnat}
\bibliography{ref}

\appendix
\clearpage

% ============================================================
\section{Per-scene file schema and loading example}
\label{app:schema}

Each released scene is organized as a self-contained directory. The required
core files are six RGB videos (\texttt{cam\_00\_rgb.mp4} through
\texttt{cam\_05\_rgb.mp4}), six metric-depth videos
(\texttt{cam\_00\_depth.mkv} through \texttt{cam\_05\_depth.mkv}), and three
JSON metadata files: \texttt{baseline.json}, \texttt{trajectory.json}, and
\texttt{\_scene\_complete.json}. RGB videos are encoded at
$1280\times1280$, $15$ fps, and $81$ frames. Depth videos use the same spatial
and temporal resolution. The released evaluator decodes depth values using the
scale specified by the dataset metadata; if a per-file scale tag is absent, the
evaluator uses the generation default of $0.1$ meters per stored unit.

\paragraph{\texttt{baseline.json}.}
This file records scene identifiers, map name, units, camera count, physical
focal length, sensor width and height, primary stereo-pair indices, adjacent
baselines, all $\binom{6}{2}=15$ pairwise baselines, and asset identifiers used
for traceability. The evaluator derives pixel focal lengths from the recorded
physical focal length and sensor dimensions as described in
Eq.~\ref{eq:intrinsics-conversion}.

\paragraph{\texttt{trajectory.json}.}
This file records per-frame six-camera poses. Its \texttt{frames} field contains
$81$ entries, one for each frame index from $0$ to $80$, and each entry stores
all six camera poses in the recorded rotation representation. Top-level fields
also record scene label, scene index, map name, character identifier, unit,
rotation representation, camera count, primary stereo pair, and camera order.

\paragraph{\texttt{\_scene\_complete.json}.}
This completion marker records the scene label, scene index, map name, and map
path. Its presence indicates that the scene passed the release audit for the
required RGB videos, depth videos, geometry metadata, trajectory metadata, and
completion marker.
\paragraph{Scene descriptions.}
Each scene also includes machine-generated scene-description metadata. The
per-scene files are \path{scene_description_qwen_dataset_level.json} and
\path{scene_description_qwen_dataset_level.txt}. The JSON file stores a
structured scene summary, including scene type, setting, primary elements,
spatial layout, lighting/materials, short and detailed captions, and tags. The
text file contains only the detailed caption and is provided for prompt-based
loaders.

Descriptions are generated once per scene from frame \(19\) of
\(\mathrm{cam}_{00}\) using Qwen/Qwen2.5-VL-3B-Instruct. A merged JSONL summary
with one record per released scene is included under \path{metadata/}. These
captions are descriptive metadata for browsing, retrieval, and prompt
construction; they are not human-verified semantic labels or frame-accurate
annotations for every camera and time step.

The snippet below loads one stereo pair at the scene's primary baseline.
Extending to all $\binom{6}{2}=15$ pairs or all $81$ frames amounts to iterating
over camera-pair indices and video frames. The released evaluation repository
contains the full loader used for reference-disparity derivation and metric
computation.

{\small
\begin{verbatim}
import json, re, av
from pathlib import Path

scene = Path("eval/MapSeenInTrain/IPD_Gaussian/AssetsvilleTown/scene_000000")
baseline = json.loads((scene / "baseline.json").read_text())
trajectory = json.loads((scene / "trajectory.json").read_text())

# The primary stereo pair is stored as camera names (e.g. "..._Cam_00")
# in the released schema; older variants used integer indices.
def cam_idx(name):
    if isinstance(name, int):
        return name
    m = re.search(r"[Cc]am_?(\d+)$", name)
    if m is None:
        raise ValueError(f"cannot parse camera index from {name!r}")
    return int(m.group(1))

pair = baseline["primary_stereo_pair"]
if isinstance(pair, dict):
    L_idx = cam_idx(pair["left_camera"])
    R_idx = cam_idx(pair["right_camera"])
else:
    L_idx, R_idx = pair

intr  = baseline["camera_intrinsics"]
f_mm  = intr["focal_length_mm"]
sw_mm = intr["sensor_width_mm"]
W = baseline.get("image_width", 1280)
fx_px = f_mm / sw_mm * W

# Realized baseline for any (L, R) pair, adjacent or not.
def baseline_cm_for(a, b, doc):
    a, b = sorted((a, b))
    for entry in doc["pairwise_pairs"]:
        ea = entry["camera_index_a"]
        eb = entry["camera_index_b"]
        if sorted((ea, eb)) == [a, b]:
            return entry["baseline_cm"]
    raise KeyError((a, b))

B_cm = baseline_cm_for(L_idx, R_idx, baseline)

def first_rgb_frame(path):
    with av.open(str(path)) as container:
        for frame in container.decode(video=0):
            return frame.to_ndarray(format="rgb24")
    raise RuntimeError(f"no rgb frame in {path}")

I_L = first_rgb_frame(scene / f"cam_{L_idx:02d}_rgb.mp4")
I_R = first_rgb_frame(scene / f"cam_{R_idx:02d}_rgb.mp4")
\end{verbatim}}
% ============================================================
\section{Realized baseline distributions}
\label{app:trainstats}

This appendix complements Table~\ref{tab:baseline-stats} by documenting the
realized baseline distributions in the released dataset. The statistics below
are computed from the final manifest used for the paper.

\begin{figure}[!h]
  \centering
  \includegraphics[width=\linewidth]{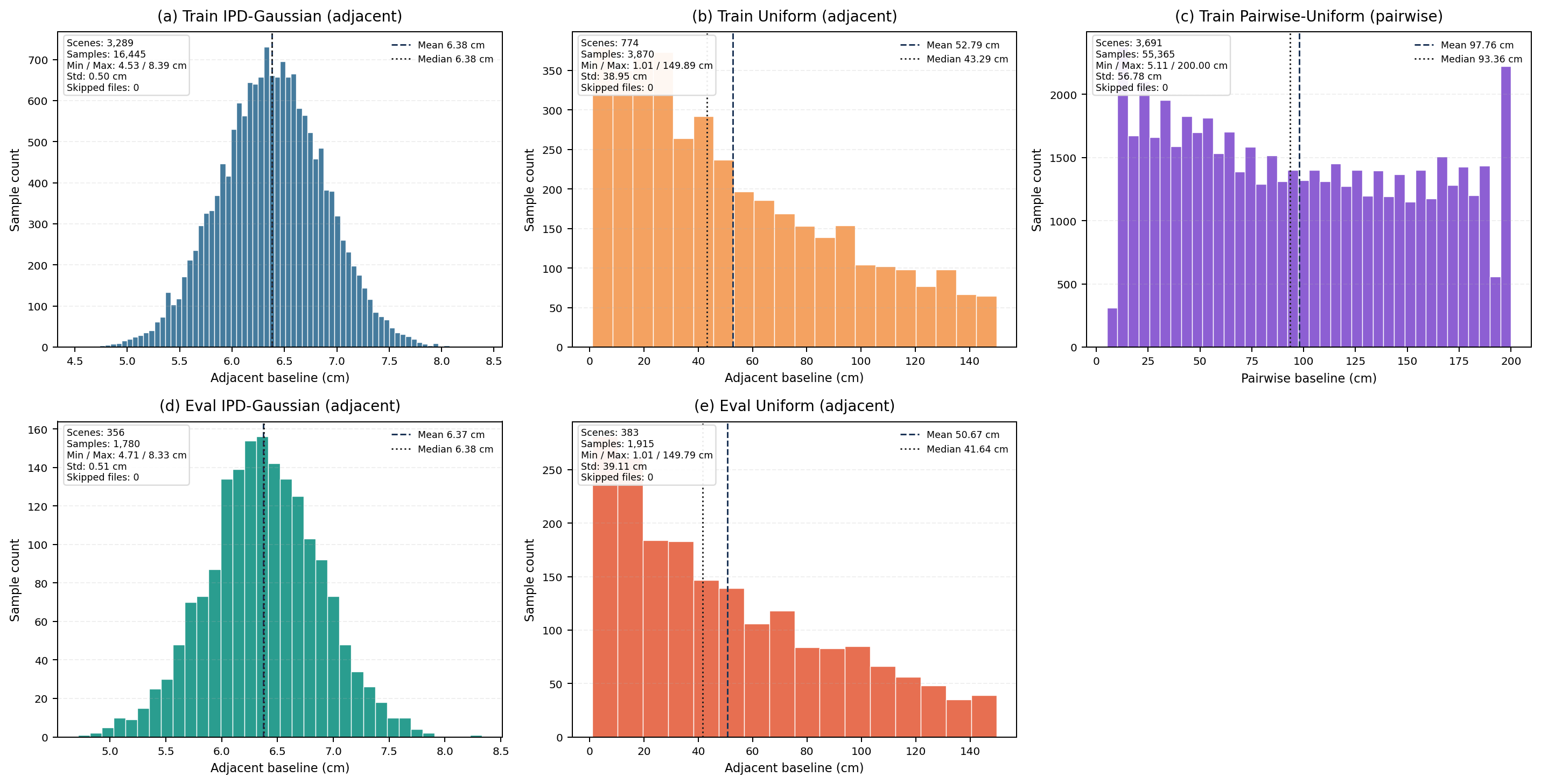}
  \caption{Realized baseline distributions in the released dataset. Panels (a), (b), (d), and (e) show adjacent-baseline distributions for \emph{IPD\_Gaussian} and \emph{Uniform} on the train and evaluation splits. Panel (c) shows the all-pairs distribution for the train-only \emph{Pairwise\_Uniform} family across all $\binom{6}{2}=15$ camera pairs. The dashed and dotted vertical lines mark the realized mean and median.}
  \label{fig:app-baseline-histograms}
\end{figure}

\paragraph{Adjacent baselines.}
The \emph{IPD\_Gaussian} family targets a truncated Gaussian centered at
$6.38$ cm with standard deviation $0.5$ cm and clamp range $[4.5,8.5]$ cm.
The released train split contains $16{,}445$ adjacent baselines in this family
(min $4.5278$ cm, max $8.3867$ cm, mean $6.3813$ cm, std $0.4998$ cm), and the
evaluation split contains $1{,}780$ adjacent baselines (min $4.7119$ cm,
max $8.3332$ cm, mean $6.3743$ cm, std $0.5052$ cm).

The \emph{Uniform} family proposes adjacent gaps from
$\mathcal{U}[1.0,150.0]$ cm. The released train split contains $3{,}870$
adjacent baselines (min $1.0113$ cm, max $149.8906$ cm, mean $52.7908$ cm,
std $38.9500$ cm, median $43.2916$ cm), and the evaluation split contains
$1{,}915$ adjacent baselines (min $1.0058$ cm, max $149.7889$ cm, mean
$50.6741$ cm, std $39.1111$ cm, median $41.6438$ cm). The realized medians are
below the theoretical median of $75.5$ cm because large adjacent gaps create
larger six-camera spans and more frequently fail collision, visibility, or
framing validation in constrained maps.

\paragraph{Train-only all-pairs coverage.}
The \emph{Pairwise\_Uniform} family is train-only and targets broad all-pairs
coverage when each six-camera scene is expanded to all $15$ camera pairs. Its
adjacent-gap statistics are reported in Table~\ref{tab:baseline-stats}. Across
all $15$ pairwise combinations, the released \emph{Pairwise\_Uniform} split
contains $55{,}365$ baselines with min $5.1117$ cm, max $199.9969$ cm, mean
$97.7615$ cm, standard deviation $56.7814$ cm, and median $93.3603$ cm. A small
low-baseline tail below the nominal $10$ cm target occurs in the release:
$150$ pairwise baselines from $30$ \texttt{UndergroundSciFi} scenes fall below
$10$ cm. We therefore describe this family as targeting broad all-pairs coverage
up to $200$ cm and report realized statistics rather than claiming that every
released all-pairs baseline lies in $[10,200]$ cm.

Users requiring a strictly uniform adjacent-baseline or all-pairs distribution
should subsample the released manifest or generate additional scenes under
modified validation thresholds.

% ============================================================
\section{Per-map scene counts}
\label{app:map-counts}

Tables~\ref{tab:app-map-counts-eval} and~\ref{tab:app-map-counts-train} report
scene counts per map and sampling family. Maps appearing in both train and
evaluation splits are listed in both tables. In the evaluation table, the subset
column indicates whether the map is also present in training (seen-map) or is
map-disjoint from training (unseen-map). Sparse cells reflect maps that passed
generation-pipeline validation for only a small number of scenes; we report this
distribution so users can construct protocols with sufficient per-map sample
sizes.

\begin{table}[!h]
  \centering
  \scriptsize
  \setlength{\tabcolsep}{4pt}
  \caption{Evaluation split: scene counts per (family, map).}
  \label{tab:app-map-counts-eval}
  \begin{tabular}{l l rr}
    \toprule
    Map & Eval subset & IPD\_Gaussian & Uniform \\
    \midrule
    \texttt{AssetsvilleTown}                  & seen-map   & 30 & 31 \\
    \texttt{Car\_Dealer}                      & seen-map   &  1 &  2 \\
    \texttt{DeepWaterStation\_DemoMapScalabilityEpic} & unseen-map & 34 & 45 \\
    \texttt{Downtown\_West}                   & seen-map   & 21 & 12 \\
    \texttt{GangnyeongieonComplex}            & seen-map   & 99 & --- \\
    \texttt{GeunjeongjeonComplex}             & seen-map   &  2 & 52 \\
    \texttt{JapaneseStyleRoom}                & seen-map   & 22 & --- \\
    \texttt{JesuhabComplex}                   & seen-map   & --- &  1 \\
    \texttt{Light\_Foliage}                   & seen-map   & 13 & 35 \\
    \texttt{Modular\_Scifi\_Mechanic\_Base}  & seen-map   &  2 & 14 \\
    \texttt{OWD\_Yucca\_Pack}                 & seen-map   & 25 & --- \\
    \texttt{ProceduralNtr\_vol2}              & seen-map   & 23 & 18 \\
    \texttt{RestaurantScene}                  & seen-map   &  2 &  1 \\
    \texttt{ROCKY\_SAND\_PACK}                & seen-map   & --- & 20 \\
    \texttt{STF\_TestMap}                     & unseen-map & 57 & 57 \\
    \texttt{Scene\_Bazaar\_Vol1}              & seen-map   & --- & 34 \\
    \texttt{Scene\_Warehouse}                 & seen-map   & --- & 30 \\
    \texttt{Stylized\_Egypt}                  & seen-map   & --- & 31 \\
    \texttt{UndergroundSciFi}                 & seen-map   & 25 & --- \\
    \midrule
    \textbf{Total}                            &            & \textbf{356} & \textbf{383} \\
    \bottomrule
  \end{tabular}
\end{table}

\begin{table}[!h]
  \centering
  \scriptsize
  \setlength{\tabcolsep}{4pt}
  \caption{Train split: scene counts per (family, map).}
  \label{tab:app-map-counts-train}
  \begin{tabular}{lrrr}
    \toprule
    Map & IPD\_Gaussian & Uniform & Pairwise\_Uniform \\
    \midrule
    \texttt{AbandonedPowerPlant}              &  10 &  27 &  30 \\
    \texttt{AssetsvilleTown}                  & 514 &   7 & 978 \\
    \texttt{Car\_Dealer}                      & 255 &  39 &   9 \\
    \texttt{Downtown\_West}                   &  62 &  12 & 120 \\
    \texttt{GangnyeongieonComplex}            & 801 &   9 & 309 \\
    \texttt{GeunjeongjeonComplex}             & 211 & 180 & 200 \\
    \texttt{JapaneseStyleRoom}                & 309 &  39 &  99 \\
    \texttt{JesuhabComplex}                   & --- &  30 & 479 \\
    \texttt{Light\_Foliage}                   &  18 &  19 & --- \\
    \texttt{Modular\_Scifi\_Mechanic\_Base}  & 523 &  27 & 132 \\
    \texttt{OWD\_Yucca\_Pack}                 &  63 &  40 &  28 \\
    \texttt{ProceduralBuildingGenerator}      & --- &  58 & --- \\
    \texttt{ProceduralNtr\_vol2}              &  83 &  22 & 517 \\
    \texttt{RestaurantScene}                  & 125 &  35 &  25 \\
    \texttt{ROCKY\_SAND\_PACK}                &  83 &  95 & --- \\
    \texttt{Scene\_Bazaar\_Vol1}              & --- &   5 & --- \\
    \texttt{Scene\_Saloon}                    &  12 & --- & --- \\
    \texttt{Scene\_UnfinishedBuilding}        &   8 &   2 &   7 \\
    \texttt{Scene\_Warehouse}                 &  50 &   7 & --- \\
    \texttt{SeyeonjeongPavilion}              & --- & --- & 698 \\
    \texttt{Stylized\_Egypt}                  & --- & 108 &  30 \\
    \texttt{UndergroundSciFi}                 & 121 & --- &  30 \\
    \texttt{UtopianCity}                      &  41 &  13 & --- \\
    \midrule
    \textbf{Total}                            & \textbf{3{,}289} & \textbf{774} & \textbf{3{,}691} \\
    \bottomrule
  \end{tabular}
\end{table}

% ============================================================
\section{Geometry derivation and projection details}
\label{app:geometry}

This appendix expands the geometry conventions summarized in
Section~\ref{sec:geometry-validation}. Each camera is modeled by an intrinsic
matrix
\begin{equation}
K =
\begin{bmatrix}
  f_x & 0 & c_x \\
  0 & f_y & c_y \\
  0 & 0 & 1
\end{bmatrix},
\end{equation}
where $f_x$ and $f_y$ are derived from the physical focal length and sensor
dimensions by Eq.~\ref{eq:intrinsics-conversion}. The principal point
$(c_x,c_y)$ defaults to the image center unless an override is recorded in
\texttt{baseline.json}.

The released evaluator treats decoded depth as camera-frame optical-axis depth.
For a source pixel $u=(x,y,1)^\top$ with valid depth $z$, the evaluator
back-projects
\begin{equation}
  X_L = z K_L^{-1} u .
\end{equation}
Given the relative pose from source camera $L$ to target camera $R$, recorded in
\texttt{trajectory.json}, the corresponding target-camera point is
\begin{equation}
  X_R = R_{RL}X_L + t_{RL}, \qquad u_R \sim K_R X_R .
\end{equation}
For the rectified lateral rig, this reduces to Eq.~\ref{eq:rectified-disparity},
up to the selected camera ordering and sign convention.

The validation audit in Table~\ref{tab:geometry-validation-main} checks
intrinsics reconstruction, pose-derived baselines, valid-depth ratio, and
non-occluded reprojection error. Non-occluded pixels are selected by comparing
projected source points against the target depth buffer and rejecting projected
pixels whose target-depth disagreement exceeds the evaluator threshold. The
released evaluation code provides the exact non-occlusion mask, depth filtering,
projection, and baseline-recovery implementation.

% ============================================================
\section{Inference conditions and dash classifications}
\label{app:interfaces}

This appendix records the inference condition used for every evaluated method:
the inputs consumed by the released checkpoint, any benchmark geometry used at
inference, the working resolution, and the resulting alignment property of the
generated right view. The tiers match Section~\ref{sec:benchmark}: G0 methods
receive target-view geometry, G1 methods receive target-camera metadata, and G2
methods receive no benchmark target-view information. 

Where a cell in the result tables is unavailable, the dash indicates one of
three cases: (i) no public code or weights were available at the time of our
evaluation, (ii) public code was available but that branch was not completed in
the evaluation run, or (iii) the evaluation was attempted but output was unusable
(e.g., decoder failure or scale collapse).

\paragraph{Unavailable and diagnostic rows.}
Dashes do not indicate zero performance. They mark evaluations that were not
available in the reported run under the released evaluator. In particular,
partial rows can occur when some metrics were computed but other branches were
not completed. Methods such as Stereo-from-Mono and StereoCrafter-Zero are
reported as diagnostics under their released interfaces rather than as strict
peers of left-to-right generators under the same input contract. Likewise,
methods omitted entirely from the quantitative tables are omitted because public
code or weights were unavailable at the time of evaluation, not because they
were judged qualitatively unimportant.

\subsection{Stereo image generation methods}
\label{app:interfaces-image}

\begin{table}[!h]
  \centering
  \scriptsize
  \setlength{\tabcolsep}{3pt}
  \caption{Inference conditions for stereo image generation methods.}
  \label{tab:app-interfaces-image}
  \begin{tabular}{l p{2.6cm} p{4cm} p{3cm} c}
    \toprule
    Method & Inputs at inference & Geometry source & Working res. & Tier \\
    \midrule
    GenStereo            & $I_L$ frames + GT disparity      & benchmark GT disparity                                & 768                & G0 \\
    StereoDiffusion      & $I_L$ + oracle disparity         & benchmark GT disparity                                & 512                & G0 \\
    ZeroStereo           & $I_L$ + GT disparity (\texttt{.npy}) & benchmark GT disparity                            & native             & G0 \\
    Stereo-from-Mono     & $I_L$ + GT-warped right          & benchmark GT disparity; constructs stereo input       & $320{\times}192$   & G0 \\
    StereoSpace          & $I_L$ + intrinsics + baseline    & manifest \texttt{baseline\_m}, \texttt{fx\_px}        & cfg. square crop  & G1 \\
    Mono2Stereo          & $I_L$                            & model-internal monocular depth; fixed viewpoint baseline 8.0, focal 0.6 & $1280{\times}800$ & G2 \\
    \bottomrule
  \end{tabular}
\end{table}

Stereo-from-Mono is included only as a diagnostic because the evaluated
checkpoint is the official stereo matching disparity network, not a single-image
right-view generator. Its output should not be interpreted as monocular stereo
generation.

\subsection{Stereo video generation methods}
\label{app:interfaces-video}

\begin{table}[!h]
  \centering
  \scriptsize
  \setlength{\tabcolsep}{3pt}
  \caption{Inference conditions for stereo video generation methods.}
  \label{tab:app-interfaces-video}
  \begin{tabular}{l p{3.0cm} p{4cm} p{3cm} c}
    \toprule
    Method & Inputs at inference & Geometry source & Working res. & Tier \\
    \midrule
    SVG                  & $\{I_L^{(t)}\}$ + GT disparity as inverse depth + scene text prompt & benchmark GT disparity & $576{\times}320$ & G0 \\
    StereoCrafter        & $\{I_L^{(t)}\}$ + GT-disparity splat & benchmark GT disparity; forward warp & $512$ & G0 \\
    ImmersePro           & $\{I_L^{(t)}\}$ & implicit model geometry & $384{\times}384$ & G2 \\
    StereoCrafter-Zero   & single $I_L$ frame + fixed prompt & internal monocular-depth prior & default & G2 \\
    StereoPilot          & $\{I_L^{(t)}\}$ + fixed prompt & implicit model geometry & $832{\times}480$, 16 fps, 81 frames & G2 \\
    \bottomrule
  \end{tabular}
\end{table}

StereoCrafter-Zero is evaluated under its single-frame-to-stereo-video
interface. It is not invoked as a left-video-conditioned paired-view converter,
which contributes to its large distributional and matchability errors.

\section{Trajectory construction and generation parameters}
\label{app:trajectory-construction}
\label{app:pipeline-params}

\begin{figure}[!h]
  \centering
  \includegraphics[width=\linewidth]{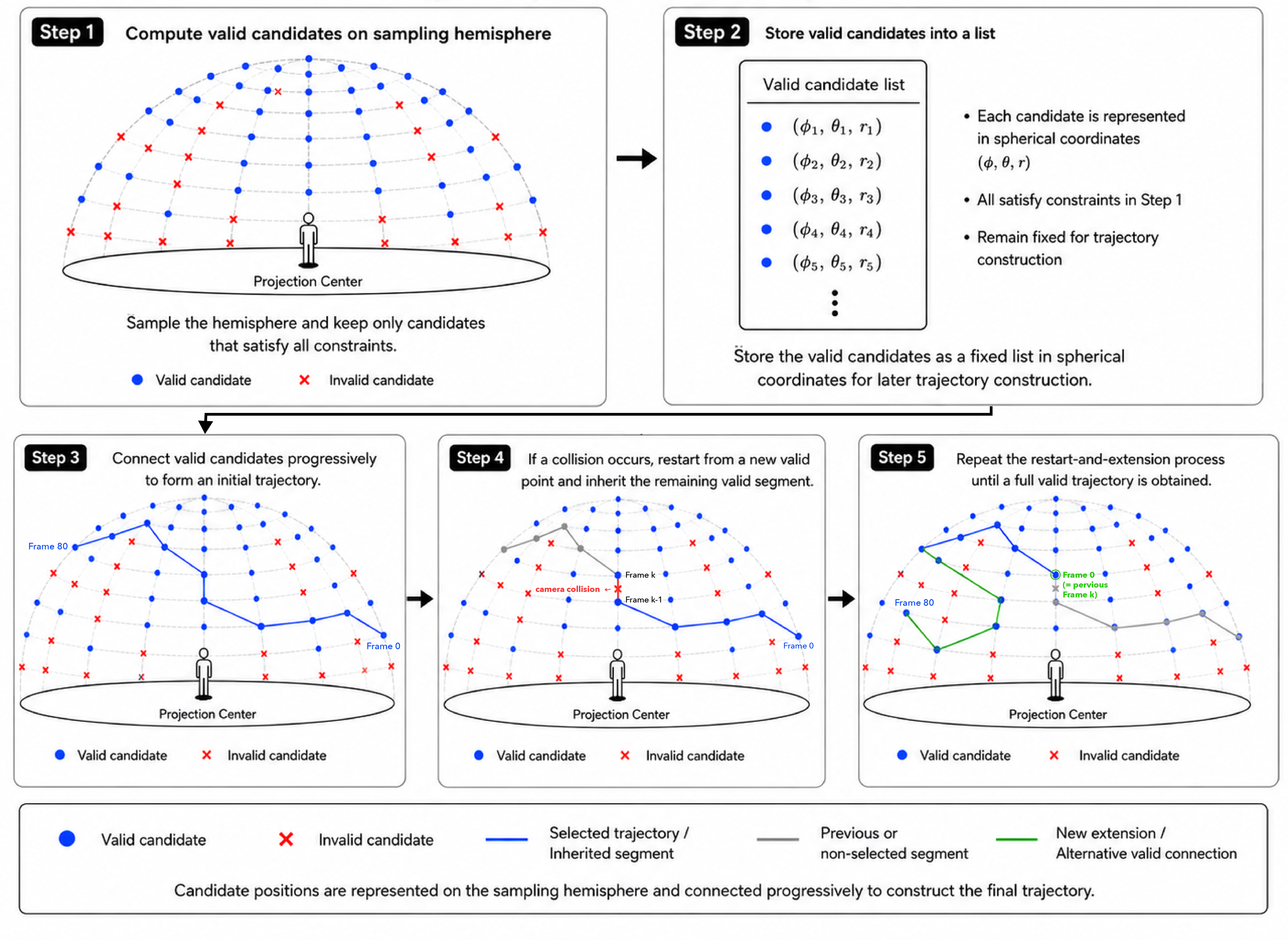}
  \caption{Trajectory construction. Candidate viewpoints are sampled around the
  subject, invalid candidates are rejected, and a complete $81$-frame trajectory
  is formed by connecting valid candidates with restart-and-splice recovery when
  a transition fails validation.}
  \label{fig:app-trajectory}
\end{figure}

The base camera trajectory is constructed over the full $81$-frame sequence.
The pipeline samples a hemisphere of candidate viewpoints around the subject and
keeps only candidates satisfying collision, visibility, framing, and motion
constraints. A frame-by-frame trajectory is progressively connected through the
valid pool; if a transition fails validation, the pipeline restarts from a valid
point and splices the remaining valid segment until a complete trajectory is
obtained. Once selected, the trajectory is materialized as a six-camera rig by
applying the sampled lateral offsets, and the dense per-frame poses are
validated before release. The full procedure is illustrated in
Figure~\ref{fig:app-trajectory}.

\paragraph{Sampling configuration.}
Per scene, focal length is sampled from $\mathcal{U}[18,85]$ mm. Sensor
dimensions are recorded in \texttt{baseline.json}. The \emph{IPD\_Gaussian}
family samples adjacent gaps from a truncated Gaussian with mean $6.38$ cm,
standard deviation $0.5$ cm, and clamp range $[4.5,8.5]$ cm. The
\emph{Uniform} family samples adjacent gaps from $\mathcal{U}[1.0,150.0]$ cm.
The train-only \emph{Pairwise\_Uniform} family targets broad all-pairs coverage
up to $200$ cm when all $15$ camera pairs are used; its realized release
statistics are reported in Appendix~\ref{app:trainstats}. Production renders
use $1280\times1280$ resolution, $81$ frames, and $15$ fps.

% ============================================================
\section{Unseen-map evaluation}
\label{app:unseen-results}

This appendix reports reference results on the map-disjoint unseen subset,
complementing the seen-map results in Section~\ref{sec:baselines}. The unseen
subset contains $193$ evaluation scenes: $91$ from \emph{IPD\_Gaussian} and
$102$ from \emph{Uniform}. These scenes are drawn from maps that do not appear
in the training split. Because public methods are evaluated off-the-shelf rather
than retrained on StereoGenBench, unseen-map results should be treated as a
generalization diagnostic rather than as the primary reference point.

\begin{table*}[!h]
  \centering
  \tiny
  \setlength{\tabcolsep}{4pt}
  \caption{Unseen-map evaluation. Branches Unseen-IPD\_Gaussian and
  Unseen-Uniform denote evaluation subsets whose maps do not appear in
  training. Metrics use the same columns and definitions as
  Table~\ref{tab:main-results}. StereoCrafter-Zero distributional metrics are
  omitted for the same single-frame-conditioned 16-frame output caveat.}
  \label{tab:app-unseen-results}
  \begin{tabular}{l c c l !{\vrule width 1pt} r r r !{\vrule width 1pt} r r r !{\vrule width 1pt} r r}
    \toprule
                                       &       &     &                       & \multicolumn{3}{c!{\vrule width 1pt}}{Reconstruction} & \multicolumn{3}{c!{\vrule width 1pt}}{Geometric correctness} & \multicolumn{2}{c}{Distributional} \\
    Method & Type & Tier & Branch & PSNR$\uparrow$ & SSIM$\uparrow$ & LPIPS$\downarrow$ & $\mathcal{E}_\mathrm{Match}\downarrow$ & P-PSNR$\uparrow$ & SD$\downarrow$ & FID$\downarrow$ & FVD$\downarrow$ \\
    \midrule
    \multicolumn{12}{l}{\emph{Tier G0 --- target-view geometry}} \\
    GenStereo                          & Image & G0 & Unseen-IPD\_Gaussian & 28.34 & 0.7713 & 0.1517 & 45.14 & 27.57 & 0.0457 &  10.94 &   77.69 \\
                                       &       &    & Unseen-Uniform       & 28.22 & 0.7560 & 0.2182 & 48.18 & 26.46 & 0.2184 &  18.83 &  329.07 \\
    StereoDiffusion                    & Image & G0 & Unseen-IPD\_Gaussian & 23.80 & 0.6193 & 0.2130 & 46.73 & 26.03 & 0.2428 &  17.57 &  153.75 \\
                                       &       &    & Unseen-Uniform       & 24.38 & 0.6369 & 0.2830 & 52.55 & 25.02 & 0.7308 &  44.57 &  762.03 \\
    ZeroStereo                         & Image & G0 & Unseen-IPD\_Gaussian & 29.11 & 0.8633 & 0.0927 & 38.73 & 29.54 & 0.0245 &   9.53 &   91.73 \\
                                       &       &    & Unseen-Uniform       & 25.38 & 0.7944 & 0.1905 & 40.21 & 26.19 & 0.1108 &  25.87 &  527.11 \\
    Stereo-from-Mono                   & Image & G0 & Unseen-IPD\_Gaussian & 25.01 & 0.7690 & 0.1271 & 39.06 & 32.02 & 0.2219 &  12.53 &   41.64 \\
                                       &       &    & Unseen-Uniform       & 25.00 & 0.6823 & 0.2123 & 40.15 & 27.78 & 0.2398 &  36.80 &  200.78 \\
    SVG                                & Video & G0 & Unseen-IPD\_Gaussian & 24.97 & 0.6726 & 0.2961 & 45.89 & 25.62 & 0.0739 &  44.65 &  123.63 \\
                                       &       &    & Unseen-Uniform       & 24.48 & 0.7455 & 0.2708 & 53.11 & 24.90 & 0.1419 &  50.25 &  288.64 \\
    StereoCrafter                      & Video & G0 & Unseen-IPD\_Gaussian & 22.76 & 0.5513 & 0.4079 & 45.98 & 23.52 & 0.1762 &  45.69 &  383.28 \\
                                       &       &    & Unseen-Uniform       & 22.06 & 0.5254 & 0.5052 & 52.51 & 22.93 & 0.6125 &  74.59 &  879.08 \\
    \midrule
    \multicolumn{12}{l}{\emph{Tier G1 --- target-camera metadata}} \\
    StereoSpace                        & Image & G1 & Unseen-IPD\_Gaussian & 20.59 & 0.5133 & 0.2589 & 46.19 & 25.15 & 0.4431 &  13.37 &   81.93 \\
                                       &       &    & Unseen-Uniform       & 22.30 & 0.5677 & 0.3113 & 52.36 & 26.12 & 0.6978 &  24.88 &  666.73 \\
    \midrule
    \multicolumn{12}{l}{\emph{Tier G2 --- unaligned monocular}} \\
    Mono2Stereo                        & Image & G2 & Unseen-IPD\_Gaussian & 19.31 & 0.4996 & 0.3140 & 52.41 & 26.14 & 2.4024  &  10.90 &   32.31 \\
                                       &       &    & Unseen-Uniform       & 21.23 & 0.5243 & 0.3718 & 58.28 & 26.98 & 28.2631 &  24.76 &   97.12 \\
    ImmersePro                         & Video & G2 & Unseen-IPD\_Gaussian & 19.08 & 0.5111 & 0.3695 & 50.97 & 25.03 & 0.9968  &   9.63 &   34.27 \\
                                       &       &    & Unseen-Uniform       & 21.01 & 0.5256 & 0.4182 & 56.32 & 26.28 & 2.9956  &  22.78 &  100.01 \\
    StereoCrafter-Zero                 & Video & G2 & Unseen-IPD\_Gaussian & 12.62 & 0.3438 & 0.7371 & 81.87 & 13.84 & 0.7943  &  ---   &  ---    \\
                                       &       &    & Unseen-Uniform       & 12.95 & 0.3402 & 0.7012 & 73.41 & 14.03 & 1.8213  &  ---   &  ---    \\
    StereoPilot                        & Video & G2 & Unseen-IPD\_Gaussian & 17.00 & 0.4437 & 0.5864 & 58.13 & 19.02 & 0.7227  &  64.65 &  443.78 \\
                                       &       &    & Unseen-Uniform       & 18.83 & 0.5029 & 0.5676 & 53.54 & 21.14 & 1.6745  &  75.09 &  375.96 \\
    \bottomrule
  \end{tabular}
\end{table*}

Across evaluated methods, $\mathcal{E}_{\mathrm{Match}}$ values on the unseen
subset are similar in magnitude to the seen subset, while FID and FVD show
larger seen--unseen gaps for several methods. This is consistent with
matchability being driven primarily by inference contract and geometry, and
distributional metrics being more sensitive to map appearance. Users should
interpret each metric's seen--unseen gap as a diagnostic of visual-prior
alignment rather than as a standalone model ranking.

% ============================================================
% ============================================================

% ============================================================
\section{Metric definitions and calibration controls}
\label{app:metric-calibration}

This appendix specifies the stereo diagnostics used in
Section~\ref{sec:benchmark} and provides evaluator controls for interpreting
their absolute scale. All frame-wise metrics are computed at \(832\times480\)
resolution after the same resizing, masking, and aggregation protocol used for
Tables~\ref{tab:main-results} and~\ref{tab:app-unseen-results}.

Let \(M_{gt}\) and \(M_{pred}\) denote the sets of left-image keypoints that
participate in accepted matches between \((I_L,I_R)\) and
\((I_L,\hat I_R)\), respectively, under the same DeDoDe
matcher~\citep{edstedt2024dedode}. The released evaluator uses DeDoDe detector
weights \texttt{L-upright}, descriptor weights \texttt{B-upright}, a keypoint
budget of \(2048\), ratio threshold \(0.9\), mutual matching, and a \(2\)-pixel
epipolar tolerance. We report the same Jaccard-style matchability error as
Eq.~\ref{eq:ematch}:
\[
\mathcal{E}_{\mathrm{Match}}
=
100\left(1 -
\frac{|M_{gt}\cap M_{pred}|}{|M_{gt}\cup M_{pred}|}
\right).
\]
The union denominator penalizes both missing matchable structures and spurious
matchable structures, so lower values indicate that the generated right view
preserves the matchable stereo structure of the rendered target pair without
introducing a substantially different set of accepted matches.

P-PSNR is a target-free stereo-consistency diagnostic computed from
\(I_L\) and \(\hat I_R\) alone. For each source patch, the evaluator searches a
horizontal disparity window and records the PSNR corresponding to the best
patch MSE. The released evaluator uses \(16\times16\) patches, stride \(16\),
and integer disparities from \(2\) to \(48\) pixels. Because P-PSNR can reward
locally similar texture even when global geometry is wrong, we report it only
as a diagnostic.

SD evaluates disparity-scale fidelity. A fixed FoundationStereo
matcher~\citep{wen2025foundationstereo} estimates disparity \(\hat D\) from
\((I_L,\hat I_R)\). On finite valid pixels, after the evaluator's outlier and
occlusion filtering, the evaluator fits
\[
D_{gt}\approx a\hat D+b
\]
and reports
\[
\mathrm{SD}=|a-1|.
\]
Thus, lower SD indicates that the generated pair induces a disparity scale
closer to the synthetic ground-truth scale. The released evaluator also records
the valid-pixel ratio and residual statistics used to interpret each fit.

\paragraph{Distributional metrics.}
FID is computed between generated right views and rendered target right views
within each evaluation branch. FVD is computed between generated and rendered
right-view video sequences. For image-generation methods evaluated on video
scenes, the image model is applied independently to each frame and FVD is
computed on the resulting frame sequence; in that case FVD should not be
interpreted as evidence of explicit temporal modeling.

Table~\ref{tab:metric-calibration} reports evaluator controls used to interpret
the nonstandard stereo diagnostics. These controls are not benchmarked methods;
they define the metric scale within StereoGenBench. The table aggregates over
the full evaluation split using the same evaluator configuration as the
reference results.

\begin{table*}[!h]
  \centering
  \small
  \setlength{\tabcolsep}{5pt}
  \caption{Metric calibration controls on the full evaluation split. All
  controls use the same evaluator configuration as
  Tables~\ref{tab:main-results} and~\ref{tab:app-unseen-results}. The
  wrong-baseline control uses the same scene rendered from camera 02 as the
  candidate right view while metrics are still computed against the target
  camera 01. The random-right control uses a right view from a different scene
  under the same evaluator aggregation protocol.}
  \label{tab:metric-calibration}
  \begin{tabular}{l r r r r r r}
    \toprule
    Control right view
    & PSNR$\uparrow$
    & SSIM$\uparrow$
    & LPIPS$\downarrow$
    & $\mathcal{E}_{\mathrm{Match}}\downarrow$
    & P-PSNR$\uparrow$
    & SD$\downarrow$ \\
    \midrule
    Rendered target \(I_R\)
      & \(\infty\) & 1.0000 & 0.0000 & 0.00 & 32.63 & 0.0209 \\
    Copied left \(I_L\)
      & 19.07 & 0.6530 & 0.1804 & 0.00 & 27.04 & 97.7503 \\
    Wrong-baseline target
      & 19.08 & 0.6531 & 0.1806 & 29.10 & 27.42 & 0.5030 \\
    Random right view
      & 9.54 & 0.2228 & 0.7711 & 85.29 & 10.46 & 0.9030 \\
    \bottomrule
  \end{tabular}
\end{table*}

The rendered-target row establishes the evaluator floor under the released
preprocessing path. The copied-left control highlights that
\(\mathcal{E}_{\mathrm{Match}}\) measures matchability rather than disparity
scale: copying \(I_L\) preserves the left-image structures used for matching,
but it produces a catastrophic SD value because it contains no target-view
stereo shift. The wrong-baseline control uses a plausible right view from the
same scene but the wrong target camera; it remains locally matchable and
texture-similar, while SD and \(\mathcal{E}_{\mathrm{Match}}\) expose the
target-camera mismatch. The random-right control provides an unmatched-scene
negative control and degrades reconstruction, matchability, and patch-level
consistency simultaneously.

\section{Generation environment, compute, and asset provenance}
\label{app:generation-environment}

The released dataset was generated on a Windows 11 workstation with an Intel
Core i7-13700K CPU, $32$ GB RAM, and an AMD Radeon RX 7900 XTX GPU with
approximately $24.5$ GB dedicated VRAM. The project uses Unreal Engine 5.5.4
(build 40574608, UE5 Release-5.5) in Development configuration, with
Movie Render Pipeline, Movie Pipeline Mask Render Pass, and PCG enabled.

Recent production batches on this workstation achieved approximately
$16$--$36$ complete scenes/hour, with a typical rate in the mid-20s
scenes/hour. Extrapolated to the full $8{,}493$-scene release, dataset
generation required on the order of a few hundred GPU-hours on a single
workstation, with a central estimate of roughly $340$ GPU-hours. This estimate
covers generation throughput and excludes manual filtering, rejected-scene
reruns, script development, and dataset upload time.

The generation project uses $15$ character FBX files and $30$ animation FBX
files imported into the Unreal project under configured content roots, together
with third-party Unreal Engine maps, textures, materials, and environment
assets. The released rendered dataset does not redistribute these third-party
source assets. The dataset license applies to the rendered RGB/depth outputs,
scene manifests, split metadata, and authored metadata. It does not relicense
third-party Unreal Engine maps, character meshes, animation FBX files, textures,
materials, or other source assets. Exact regeneration or extension with the
same source assets requires users to obtain compatible assets under their
original license terms.

\section{Baseline-stratified Uniform-branch diagnostics}
\label{app:baseline-bins}

Because calibrated baseline is the main controlled axis of StereoGenBench, the
seen-map \emph{Uniform} branch is reported in realized-baseline bins. The table
below assesses whether reconstruction quality, matchability, and disparity
scale degrade as realized baseline increases. Entries are mean \(\pm\) 95\%
confidence interval over scene-level sample means. \(N_s\) is the number of
evaluated scenes or sequences in the bin, and \(N_f\) is the total number of
evaluated frames. The table reports frame-wise metrics from
Table~\ref{tab:main-results}; branch-level distributional metrics require
separate per-bin feature-statistic runs and are not averaged from frame-wise
sample metrics.

\begin{table*}[!h]
  \centering
  \tiny
  \setlength{\tabcolsep}{3pt}
  \caption{Baseline-stratified diagnostics for the \emph{Uniform} branch.
  The five bins cover the same 281 seen-map \emph{Uniform} scenes used for the
  \emph{Uniform} rows in Table~\ref{tab:main-results}. Confidence intervals are
  computed over scene-level sample means.}
  \label{tab:app-baseline-bins}
  \begin{tabular}{l l r r r r r r r r}
    \toprule
    Method / tier & Baseline bin (cm) & \(N_s\) & \(N_f\) &
    PSNR$\uparrow$ & SSIM$\uparrow$ & LPIPS$\downarrow$ &
    $\mathcal{E}_{\mathrm{Match}}\downarrow$ & P-PSNR$\uparrow$ & SD$\downarrow$ \\
    \midrule
    \multicolumn{10}{l}{\emph{Tier G0 --- target-view geometry}} \\
    GenStereo / G0 & $[1,10)$ & 52 & 4212 & 31.35$\pm$0.90 & 0.8810$\pm$0.0179 & 0.0882$\pm$0.0107 & 33.82$\pm$1.56 & 31.53$\pm$0.95 & 0.0484$\pm$0.0156 \\
    GenStereo / G0 & $[10,30)$ & 65 & 5265 & 29.36$\pm$0.86 & 0.8592$\pm$0.0195 & 0.0948$\pm$0.0103 & 39.79$\pm$2.42 & 29.04$\pm$1.09 & 0.0405$\pm$0.0192 \\
    GenStereo / G0 & $[30,60)$ & 70 & 5670 & 26.82$\pm$0.87 & 0.8219$\pm$0.0225 & 0.1198$\pm$0.0123 & 43.08$\pm$2.04 & 23.24$\pm$0.96 & 0.0290$\pm$0.0161 \\
    GenStereo / G0 & $[60,100)$ & 55 & 4455 & 24.76$\pm$0.87 & 0.7881$\pm$0.0298 & 0.1369$\pm$0.0156 & 50.44$\pm$2.58 & 19.96$\pm$0.79 & 0.0448$\pm$0.0203 \\
    GenStereo / G0 & $[100,150]$ & 39 & 3159 & 23.31$\pm$1.10 & 0.7667$\pm$0.0374 & 0.1635$\pm$0.0201 & 56.19$\pm$2.67 & 18.62$\pm$1.04 & 0.0946$\pm$0.0506 \\
    \addlinespace

    StereoDiffusion / G0 & $[1,10)$ & 52 & 4212 & 25.56$\pm$0.98 & 0.7514$\pm$0.0299 & 0.1267$\pm$0.0113 & 36.10$\pm$1.50 & 28.00$\pm$0.97 & 0.7324$\pm$0.2780 \\
    StereoDiffusion / G0 & $[10,30)$ & 65 & 5265 & 24.33$\pm$0.82 & 0.7410$\pm$0.0245 & 0.1498$\pm$0.0112 & 42.06$\pm$2.29 & 26.62$\pm$0.94 & 0.1630$\pm$0.0393 \\
    StereoDiffusion / G0 & $[30,60)$ & 70 & 5670 & 22.56$\pm$0.76 & 0.6863$\pm$0.0289 & 0.1904$\pm$0.0138 & 45.43$\pm$2.06 & 22.56$\pm$0.86 & 0.0712$\pm$0.0225 \\
    StereoDiffusion / G0 & $[60,100)$ & 55 & 4455 & 20.82$\pm$0.85 & 0.6415$\pm$0.0405 & 0.2374$\pm$0.0228 & 52.69$\pm$2.47 & 19.50$\pm$0.79 & 0.0765$\pm$0.0370 \\
    StereoDiffusion / G0 & $[100,150]$ & 39 & 3159 & 19.53$\pm$0.93 & 0.6368$\pm$0.0492 & 0.2854$\pm$0.0295 & 57.40$\pm$2.88 & 18.14$\pm$0.98 & 0.0931$\pm$0.0437 \\
    \addlinespace

    ZeroStereo / G0 & $[1,10)$ & 52 & 4212 & 35.50$\pm$0.89 & 0.9575$\pm$0.0072 & 0.0287$\pm$0.0071 & 28.49$\pm$1.97 & 35.02$\pm$1.07 & 0.0190$\pm$0.0089 \\
    ZeroStereo / G0 & $[10,30)$ & 65 & 5265 & 29.44$\pm$1.20 & 0.9062$\pm$0.0204 & 0.0787$\pm$0.0171 & 33.61$\pm$2.58 & 30.67$\pm$1.36 & 0.0193$\pm$0.0110 \\
    ZeroStereo / G0 & $[30,60)$ & 70 & 5670 & 24.17$\pm$1.04 & 0.8666$\pm$0.0210 & 0.1156$\pm$0.0161 & 34.27$\pm$2.28 & 23.12$\pm$1.02 & 0.0196$\pm$0.0166 \\
    ZeroStereo / G0 & $[60,100)$ & 55 & 4455 & 21.11$\pm$1.19 & 0.7944$\pm$0.0342 & 0.1733$\pm$0.0271 & 41.52$\pm$3.35 & 19.12$\pm$0.78 & 0.0398$\pm$0.0228 \\
    ZeroStereo / G0 & $[100,150]$ & 39 & 3159 & 18.18$\pm$1.11 & 0.7388$\pm$0.0339 & 0.2388$\pm$0.0272 & 46.94$\pm$2.85 & 17.39$\pm$1.02 & 0.0681$\pm$0.0375 \\
    \addlinespace

    Stereo-from-Mono / G0 & $[1,10)$ & 52 & 4212 & 31.26$\pm$1.03 & 0.9288$\pm$0.0113 & 0.0284$\pm$0.0047 & 30.77$\pm$1.51 & 35.34$\pm$1.19 & 0.0707$\pm$0.0169 \\
    Stereo-from-Mono / G0 & $[10,30)$ & 65 & 5265 & 25.63$\pm$1.06 & 0.8563$\pm$0.0253 & 0.0781$\pm$0.0135 & 36.70$\pm$2.39 & 32.16$\pm$1.37 & 0.0596$\pm$0.0196 \\
    Stereo-from-Mono / G0 & $[30,60)$ & 70 & 5670 & 21.24$\pm$0.88 & 0.7441$\pm$0.0323 & 0.1522$\pm$0.0187 & 39.98$\pm$2.28 & 24.86$\pm$1.15 & 0.1106$\pm$0.0323 \\
    Stereo-from-Mono / G0 & $[60,100)$ & 55 & 4455 & 19.17$\pm$0.88 & 0.6411$\pm$0.0516 & 0.2301$\pm$0.0302 & 47.76$\pm$3.01 & 20.46$\pm$0.86 & 0.1632$\pm$0.0404 \\
    Stereo-from-Mono / G0 & $[100,150]$ & 39 & 3159 & 17.64$\pm$1.12 & 0.5834$\pm$0.0643 & 0.3073$\pm$0.0439 & 56.95$\pm$3.63 & 19.03$\pm$1.15 & 0.2067$\pm$0.0612 \\
    \addlinespace

    SVG / G0 & $[1,10)$ & 52 & 4212 & 28.53$\pm$0.92 & 0.8165$\pm$0.0252 & 0.1492$\pm$0.0161 & 36.11$\pm$1.40 & 29.27$\pm$1.01 & 0.1013$\pm$0.0309 \\
    SVG / G0 & $[10,30)$ & 65 & 5265 & 25.99$\pm$0.93 & 0.7736$\pm$0.0287 & 0.1911$\pm$0.0219 & 40.50$\pm$1.72 & 27.23$\pm$1.07 & 0.0843$\pm$0.0469 \\
    SVG / G0 & $[30,60)$ & 70 & 5670 & 22.31$\pm$0.83 & 0.7282$\pm$0.0292 & 0.2326$\pm$0.0223 & 46.22$\pm$1.99 & 22.40$\pm$0.86 & 0.0594$\pm$0.0389 \\
    SVG / G0 & $[60,100)$ & 55 & 4455 & 20.44$\pm$1.00 & 0.6649$\pm$0.0397 & 0.2826$\pm$0.0330 & 53.49$\pm$2.53 & 18.88$\pm$0.74 & 0.1206$\pm$0.0942 \\
    SVG / G0 & $[100,150]$ & 39 & 3159 & 18.49$\pm$1.10 & 0.6533$\pm$0.0442 & 0.3229$\pm$0.0331 & 59.54$\pm$2.49 & 17.24$\pm$1.01 & 0.1509$\pm$0.0918 \\
    \addlinespace

    StereoCrafter / G0 & $[1,10)$ & 52 & 4212 & 25.78$\pm$0.66 & 0.7444$\pm$0.0268 & 0.2522$\pm$0.0231 & 35.02$\pm$1.54 & 26.84$\pm$0.68 & 0.2872$\pm$0.0678 \\
    StereoCrafter / G0 & $[10,30)$ & 65 & 5265 & 23.96$\pm$0.60 & 0.7211$\pm$0.0280 & 0.2753$\pm$0.0207 & 40.64$\pm$2.05 & 25.14$\pm$0.69 & 0.1695$\pm$0.0523 \\
    StereoCrafter / G0 & $[30,60)$ & 70 & 5670 & 21.38$\pm$0.64 & 0.6427$\pm$0.0275 & 0.3311$\pm$0.0217 & 46.76$\pm$2.35 & 21.47$\pm$0.73 & 0.1061$\pm$0.0415 \\
    StereoCrafter / G0 & $[60,100)$ & 55 & 4455 & 19.33$\pm$0.73 & 0.5914$\pm$0.0359 & 0.3626$\pm$0.0263 & 54.14$\pm$3.38 & 18.73$\pm$0.78 & 0.1365$\pm$0.0793 \\
    StereoCrafter / G0 & $[100,150]$ & 39 & 3159 & 18.16$\pm$1.14 & 0.5606$\pm$0.0424 & 0.4021$\pm$0.0294 & 57.09$\pm$4.81 & 17.31$\pm$1.10 & 0.1836$\pm$0.0804 \\

    \midrule
    \multicolumn{10}{l}{\emph{Tier G1 --- target-camera metadata}} \\
    StereoSpace / G1 & $[1,10)$ & 52 & 4212 & 21.69$\pm$1.04 & 0.6458$\pm$0.0394 & 0.1722$\pm$0.0154 & 34.30$\pm$1.57 & 30.43$\pm$1.14 & 0.6272$\pm$0.0489 \\
    StereoSpace / G1 & $[10,30)$ & 65 & 5265 & 22.37$\pm$0.80 & 0.6871$\pm$0.0300 & 0.1577$\pm$0.0129 & 42.24$\pm$2.81 & 27.88$\pm$0.94 & 0.3252$\pm$0.0363 \\
    StereoSpace / G1 & $[30,60)$ & 70 & 5670 & 19.65$\pm$0.74 & 0.5810$\pm$0.0388 & 0.2295$\pm$0.0216 & 46.47$\pm$2.26 & 23.44$\pm$1.06 & 0.3369$\pm$0.0472 \\
    StereoSpace / G1 & $[60,100)$ & 55 & 4455 & 17.22$\pm$0.87 & 0.4795$\pm$0.0551 & 0.3206$\pm$0.0341 & 54.06$\pm$3.10 & 19.31$\pm$1.09 & 0.3718$\pm$0.0551 \\
    StereoSpace / G1 & $[100,150]$ & 39 & 3159 & 16.22$\pm$1.03 & 0.4694$\pm$0.0595 & 0.3921$\pm$0.0356 & 60.02$\pm$2.45 & 17.25$\pm$1.20 & 0.3472$\pm$0.0443 \\

    \midrule
    \multicolumn{10}{l}{\emph{Tier G2 --- unaligned monocular}} \\
    Mono2Stereo / G2 & $[1,10)$ & 52 & 4212 & 22.85$\pm$0.94 & 0.6835$\pm$0.0368 & 0.1231$\pm$0.0130 & 37.29$\pm$2.03 & 26.37$\pm$1.00 & 2.6279$\pm$0.8333 \\
    Mono2Stereo / G2 & $[10,30)$ & 65 & 5265 & 19.62$\pm$0.81 & 0.6186$\pm$0.0346 & 0.2383$\pm$0.0192 & 46.13$\pm$2.28 & 25.87$\pm$0.81 & 10.2501$\pm$3.6597 \\
    Mono2Stereo / G2 & $[30,60)$ & 70 & 5670 & 17.22$\pm$0.71 & 0.5026$\pm$0.0410 & 0.3458$\pm$0.0253 & 53.76$\pm$2.10 & 25.06$\pm$0.88 & 17.0041$\pm$6.7844 \\
    Mono2Stereo / G2 & $[60,100)$ & 55 & 4455 & 15.96$\pm$0.72 & 0.4370$\pm$0.0540 & 0.3951$\pm$0.0275 & 61.93$\pm$2.27 & 24.21$\pm$0.91 & 24.3151$\pm$9.8736 \\
    Mono2Stereo / G2 & $[100,150]$ & 39 & 3159 & 15.40$\pm$0.93 & 0.4493$\pm$0.0631 & 0.4277$\pm$0.0295 & 67.56$\pm$2.06 & 24.30$\pm$1.24 & 50.8491$\pm$21.1160 \\
    \addlinespace

    ImmersePro / G2 & $[1,10)$ & 52 & 4212 & 22.10$\pm$0.91 & 0.6926$\pm$0.0347 & 0.1733$\pm$0.0179 & 35.63$\pm$2.22 & 25.12$\pm$0.94 & 0.9145$\pm$0.0637 \\
    ImmersePro / G2 & $[10,30)$ & 65 & 5265 & 19.44$\pm$0.80 & 0.6221$\pm$0.0346 & 0.2710$\pm$0.0197 & 44.40$\pm$2.47 & 24.88$\pm$0.82 & 1.1411$\pm$0.2922 \\
    ImmersePro / G2 & $[30,60)$ & 70 & 5670 & 17.11$\pm$0.70 & 0.5026$\pm$0.0412 & 0.3703$\pm$0.0256 & 51.96$\pm$2.20 & 23.82$\pm$0.71 & 2.1561$\pm$0.8206 \\
    ImmersePro / G2 & $[60,100)$ & 55 & 4455 & 15.91$\pm$0.72 & 0.4385$\pm$0.0541 & 0.4152$\pm$0.0287 & 60.24$\pm$2.25 & 23.24$\pm$0.77 & 2.3669$\pm$1.2110 \\
    ImmersePro / G2 & $[100,150]$ & 39 & 3159 & 15.37$\pm$0.93 & 0.4471$\pm$0.0631 & 0.4476$\pm$0.0312 & 66.58$\pm$2.13 & 23.14$\pm$1.02 & 3.5322$\pm$2.0265 \\
    \addlinespace

    StereoCrafter-Zero / G2 & $[1,10)$ & 52 & 832 & 12.49$\pm$0.42 & 0.4325$\pm$0.0377 & 0.6407$\pm$0.0218 & 59.86$\pm$5.56 & 13.65$\pm$0.45 & 0.9163$\pm$0.0154 \\
    StereoCrafter-Zero / G2 & $[10,30)$ & 65 & 1040 & 12.38$\pm$0.39 & 0.4262$\pm$0.0309 & 0.6605$\pm$0.0177 & 66.16$\pm$4.38 & 13.57$\pm$0.41 & 0.8371$\pm$0.1709 \\
    StereoCrafter-Zero / G2 & $[30,60)$ & 70 & 1120 & 12.19$\pm$0.40 & 0.3795$\pm$0.0358 & 0.6670$\pm$0.0156 & 65.30$\pm$4.58 & 13.51$\pm$0.42 & 0.7574$\pm$0.2541 \\
    StereoCrafter-Zero / G2 & $[60,100)$ & 55 & 880 & 12.20$\pm$0.49 & 0.3658$\pm$0.0437 & 0.6751$\pm$0.0196 & 75.33$\pm$4.40 & 13.53$\pm$0.51 & 0.6913$\pm$0.1927 \\
    StereoCrafter-Zero / G2 & $[100,150]$ & 39 & 624 & 11.44$\pm$0.41 & 0.3530$\pm$0.0495 & 0.6773$\pm$0.0235 & 79.23$\pm$4.83 & 12.78$\pm$0.47 & 1.1045$\pm$0.4580 \\
    \addlinespace

    StereoPilot / G2 & $[1,10)$ & 52 & 4212 & 16.08$\pm$0.79 & 0.4943$\pm$0.0415 & 0.4866$\pm$0.0244 & 34.75$\pm$4.56 & 17.99$\pm$0.83 & 0.8811$\pm$0.0363 \\
    StereoPilot / G2 & $[10,30)$ & 65 & 5265 & 15.28$\pm$0.72 & 0.4923$\pm$0.0356 & 0.5186$\pm$0.0168 & 41.81$\pm$4.25 & 17.20$\pm$0.76 & 0.8324$\pm$0.1177 \\
    StereoPilot / G2 & $[30,60)$ & 70 & 5670 & 14.59$\pm$0.66 & 0.4241$\pm$0.0393 & 0.5445$\pm$0.0197 & 50.32$\pm$4.36 & 17.02$\pm$0.70 & 0.7739$\pm$0.0909 \\
    StereoPilot / G2 & $[60,100)$ & 55 & 4455 & 13.60$\pm$0.71 & 0.3702$\pm$0.0490 & 0.5626$\pm$0.0219 & 60.37$\pm$5.65 & 16.24$\pm$0.81 & 0.8320$\pm$0.1700 \\
    StereoPilot / G2 & $[100,150]$ & 39 & 3159 & 13.48$\pm$0.74 & 0.3960$\pm$0.0556 & 0.5722$\pm$0.0265 & 56.71$\pm$8.23 & 16.31$\pm$0.88 & 1.0301$\pm$0.2792 \\
    \bottomrule
  \end{tabular}
\end{table*}

% ============================================================

% ============================================================

% ============================================================
\section{Hosting, license, Responsible AI, and reproducibility}
\label{app:hosting}

The dataset, inspection sample, generation code, and evaluation code are
available online:
\begin{itemize}
  \item \textbf{Full dataset:}
  \href{https://huggingface.co/datasets/stereo-dataset/stereo-dataset}
  {\texttt{stereo-dataset/stereo-dataset}}.

  \item \textbf{Inspection sample:}
  \href{https://huggingface.co/datasets/stereo-dataset/stereo-dataset-small-sample-2gb}
  {\texttt{stereo-dataset/stereo-dataset-small-sample-2gb}}.

  \item \textbf{Generation code:}
  \href{https://anonymous.4open.science/r/submission-artifact-2026a-4B74}
  {\texttt{submission-artifact-2026a-4B74}}.

  \item \textbf{Evaluation code:}
  \href{https://anonymous.4open.science/r/submission-artifact-2026a-4B74}
  {\texttt{submission-artifact-2026a-4B74}}.
\end{itemize}
The rendered RGB videos, rendered metric-depth videos, scene manifests,
per-scene metadata, split metadata, and Croissant metadata are released under
CC-BY-4.0. The generation code, evaluation code, helper scripts, and
configuration files authored for StereoGenBench are released under the MIT
License. These licenses apply only to artifacts authored and released by the
StereoGenBench authors. Source Unreal Engine maps, character meshes, animation
FBX files, textures, materials, and other third-party source assets are not
redistributed under CC-BY-4.0 or MIT and retain their original license terms.
Users do not need those source assets to evaluate methods on the released
rendered dataset; exact regeneration or extension with the same maps,
characters, and animations requires obtaining compatible source assets and
complying with their original licenses.

A Croissant metadata file is published alongside the dataset and validates
against the official Croissant validator for the submitted artifact snapshot.
It records the machine-readable dataset structure and Responsible AI fields,
including intended uses, non-recommended uses, data limitations, data biases,
personal or sensitive information, social impact, generation process, and
preprocessing.

StereoGenBench contains no real personal data. All human figures are synthetic
avatars from a limited asset pool. The intended uses are evaluation and
development of stereo generation, stereo geometry, view synthesis, and
depth-related methods under controlled camera conditions. Non-recommended uses
include surveillance, biometric identification, identity inference, demographic
analysis, human behavior recognition, and impersonation-oriented synthetic-media
training.

Exact byte-level regeneration of the released dataset requires the same Unreal
Engine version, project configuration, random seeds, source assets, GPU/driver
environment, and Movie Render Queue settings. Functional extension of the
pipeline can be performed with compatible user-provided maps, characters, and
animations, but the resulting distribution may differ from the released dataset.

% ============================================================
\section{Other tasks supported by the released metadata}
\label{app:sideline-tasks}

Although this paper evaluates right-view generation, the released
metadata supports additional geometry-aware tasks. Any of the
$\binom{6}{2}=15$ view pairs from a scene can be used for stereo matching or
right-view synthesis. The six per-camera depth streams support monocular depth
evaluation under varying focal lengths. The synchronized six-camera poses
support multi-view depth refinement and view-synthesis protocols. Because the
six cameras form a lateral rig rather than a surround-view capture, the data is
best suited to baseline-controlled stereo and narrow-baseline multi-view
refinement rather than unrestricted 3D reconstruction.

% \newpage

% \input{checklist.tex}

\end{document}